\documentclass[1p,times]{elsarticle}

\usepackage{amssymb}

\usepackage{lineno}

\usepackage{natbib}
\usepackage[fleqn]{amsmath}
\usepackage{amsthm}
\usepackage{geometry}
\usepackage{fleqn}
\usepackage{xspace}
\usepackage[ruled]{algorithm2e}
\usepackage{color}

\usepackage{times}
\usepackage{soul}
\usepackage{url}
\usepackage[utf8]{inputenc}
\usepackage{graphicx}

\journal{arxiv.org}

\usepackage{xcolor}

\newtheorem{theorem}{Theorem}[section]
\newtheorem{example}[theorem]{Example}

\newtheorem{lemma}[theorem]{Lemma}

\newtheorem{corollary}[theorem]{Corollary}

\newcommand{\dls}{\textsc{Dls}\xspace}
\newcommand{\dlsstar}{\textsc{Dls}$^*$\xspace}
\newcommand{\snc}[1]{\ensuremath{\mathit{snc}\big(#1\big)}\xspace}
\newcommand{\wsc}[1]{\ensuremath{\mathit{wsc}\big(#1\big)}\xspace}
\newcommand{\expr}{\mathit{expr}}
\newcommand{\forget}{\ensuremath{\mathit{forget}}}

\newcommand{\lfp}[2]{\ensuremath{\mathrm{Lfp}\,#1.\big(#2\big)}\xspace}
\newcommand{\gfp}[2]{\ensuremath{\mathrm{Gfp}\,#1.\big(#2\big)}\xspace}

\newcommand{\proplog}{\ensuremath{\mathcal{L}_0}\xspace}
\newcommand{\propvar}{\ensuremath{\mathcal{V}_0}\xspace}
\newcommand{\folog}{\ensuremath{\mathcal{L}_1}\xspace}
\newcommand{\fovar}{\ensuremath{\mathcal{V}_1}\xspace}
\newcommand{\solog}{\ensuremath{\mathcal{L}_2}\xspace}
\newcommand{\sovar}{\ensuremath{\mathcal{V}_2}\xspace}

\newcommand{\nforget}[1]{\ensuremath{F^{NC}\big(#1\big)}\xspace}
\newcommand{\sforget}[1]{\ensuremath{F^{SC}\big(#1\big)}\xspace}

\newcommand{\calp}{{\ensuremath{\cal P}}\xspace}
\newcommand{\calr}{{\ensuremath{\cal R}}\xspace}

\newcommand{\prob}[4]{%
	\ensuremath{\calp_{#1}^{#2}\big(#3\ifthenelse{ \equal{#4}{} }{\big)}{\mid#4\big)}
		}
	}

\def\implies{\rightarrow}

\def\lneg{\neg}
\def\true{\ensuremath{\mathbb{T}}\xspace}
\def\false{\ensuremath{\mathbb{F}}\xspace}

\def\defeq{\stackrel{\mathrm{def}}{=}}
\def\defequiv{\stackrel{\mathrm{def}}{\equiv}}

\def\done{\hspace*{\fill}
    \mbox{\hspace*{\fill}\ensuremath{\Box}}}

\def\dlsstar{\mbox{\sc Dls*}\xspace}

\begin{document}

 \begin{frontmatter}

 \title{Dual Forgetting Operators in the Context of Weakest Sufficient and Strongest Necessary Conditions }

 \author{
  Patrick Doherty
  \fnref{fn2,fn1}}
 \ead{patrick.doherty@liu.se}
 
 \address[fn1]{Department of Computer and Information Science \\
  Link\"{o}ping University,
  SE-581 83 Link\"oping, Sweden}
 \address[fn2]{Faculty of Informatics \\ Mahasarakham University, Mahasarakham, Thailand}
 
 \author{Andrzej Sza{\l}as\corref{cor1}%
 \fnref{fn1,fn3}}
 \ead{andrzej.szalas@liu.se}
 \address[fn3]{Institute of Informatics, University of Warsaw\\ Banacha 2, 02-097 Warsaw, Poland}

 \vspace*{-.5pc}

 \cortext[cor1]{Corresponding author. Full postal address: please use Link\"{o}ping University.}
 \fntext[fn1]{The first author has been partially supported by a grant from the ELLIIT Network Organization for Information and Communication Technology, Sweden,  in addition to  a research grant from Mahasarakham University, Thailand.}
 \fntext[fn3]{The second author has been supported by grant 2017/27/B/ST6/02018 of the National Science Centre Poland.}

 \begin{abstract}

 \textit{Forgetting} is an important concept in knowledge representation and automated reasoning with widespread applications across a number of disciplines. A standard forgetting operator, characterized in~\cite{Lin94forgetit} in terms of model-theoretic semantics and primarily focusing on the propositional case, opened up a new research subarea. In this paper, a new operator called \textit{weak forgetting}, dual to standard forgetting, is introduced and both together are shown to offer a new more uniform perspective on forgetting operators in general. Both the weak and standard forgetting operators are characterized in terms of entailment and inference, rather than a model theoretic semantics. This naturally leads to a useful algorithmic perspective based on quantifier elimination and the use of Ackermman's Lemma and its fixpoint generalization. The strong formal relationship between standard forgetting and strongest necessary conditions and weak forgetting and weakest sufficient conditions is also characterized quite naturally through the entailment-based, inferential perspective used. The framework used to characterize the dual forgetting operators is also generalized to the first-order case and includes useful algorithms for computing first-order forgetting operators in special cases. Practical examples are also included to show the importance of both weak and standard forgetting in modeling and representation.
 \end{abstract}

 
 \end{frontmatter}

\section{Introduction and Motivation}\label{sec:intro}

From a knowledge representation and automated reasoning perspective, \textit{remembering} is essentially what an agent system does when adding new logical statements to a knowledge or belief base. Remembering, in this context, is a powerful way of modeling and is the basis for decision making in many agent systems. On the surface, remembering appears to be straightforward, simply add a new statement to a knowledge or belief base. But what if one wants to retain consistency or some other property of the knowledge or belief base upon assertion of additional statements? The knowledge or belief base would then need to be modified in various ways. Then the problem becomes more complex and leads to different subareas in Knowledge Representation, such as belief revision~\cite{AGM85,agm25years,Peppas08} or research with consistency preserving operators~\cite{AGM85,LangM10}. 

This setting also leads naturally to the dual concept of \textit{forgetting}. Given a knowledge or belief base, what does it mean to forget parts of it permanently, or temporarily for reasons of expedience? Here, on the surface also, forgetting appears to be straightforward, simply remove a statement from a~knowledge or belief base. But, as in the case of remembering, there is a great deal of subtlety and choice  concerning why and how one might remove a statement, or parts of statements from a~knowledge or belief base.

The spectrum between explicit remembering and explicit forgetting and the operators that would be needed for specifying the different degrees in between, offer a complex set of research topics in Knowledge Representation. Lin and Reiter~\cite{Lin94forgetit} opened up a new subarea of Knowledge Representation with the introduction of a (\textit{standard}) forgetting operator applied to a knowledge or belief base. The forgetting operator is specified in terms of model-theoretic semantical criteria, as are its properties. The focus in their work is primarily propositional, but there is consideration of the first-order case. The basic question asked and answered is ``what does it mean to forget certain concepts (propositional variables) in a knowledge base and how does this influence entailment of formulas in that knowledge base?''.

This context is the starting point for this paper.
Here the interest is in exploring whether there are other well-behaved forgetting operators in the spectrum discussed above and how they may relate to the original standard forgetting operator. Such operators should also be useful representationally and pragmatically. 

Consider the following motivating example where the need for an additional forgeting operator is considered. Let $lt$ and $lp$ stand for ``low temperature'' and ``low pressure'', respectively. Assume we are modeling a physical system and want to maintain the property:
\begin{equation}\label{eq:maintain}
lt\lor lp.
\end{equation}
Consider a situation when a temperature sensor associated with the system is broken and we receive no meaningful information about $lt$. To adapt the model for this situation, we would then want to temporarily forget $lt$. According to standard approaches of forgetting, this would result in the second-order formula~\cite{Lin94forgetit}:
\begin{equation}\label{eq:ltlptrue}
    \exists\, lt\,\big(lt\lor lp\big).
\end{equation}
Formula~\eqref{eq:ltlptrue} is equivalent to \textit{true}, so this would leave us empty-handed when reasoning about maintaining~\eqref{eq:maintain} with the associated changes in the system. No additional consequences of the change in the system can be derived. This happens since standard forgetting has the property of preserving entailment (see Proposition~10, point~2 in~\cite{Lin94forgetit}), where one is interested in what a given theory entails, i.e., in  the necessary conditions of the theory. On the other hand, in this situation one would expect that $lp$ itself should still be maintained, as it is a~{\em sufficient} condition for~\eqref{eq:maintain}. However, this weaker from of reasoning is not covered by standard forgetting.

Although this is a simple example, it allows us to target what this paper is about. We are interested in this weaker form of reasoning associated with forgetting and its relation to the stronger standard form of reasoning with forgetting and how these two forms of forgetting can be used in various applications. 

The original contributions of the paper include:
\begin{itemize}
    \item complementing the standard forgetting operator with a~new one, the \textit{weak} forgetting operator, that is dual to standard forgetting, useful in applications and, surprisingly, not explicitly considered in the literature so far;\footnote{In the paper, we call the new operator the \textit{weak} forgetting operator and the standard operator found in the literature, either the \textit{standard} or the \textit{strong} forgetting operator.}
    \item specifying forgetting operators in a general, principled framework that is directly related to entailment and inference rather than through a semantic construction of model equivalence as is typically used in defining standard forgetting (see, e.g.,~\cite[Definition 1]{Lin94forgetit} and other related work);
    \item the formal framework introduced shows the strong dual relationship between standard forgetting and strongest necessary conditions, and weak forgetting and weakest sufficient conditions. This relationship follows naturally from the entailment-based, inferential perspective used; 
    \item a computational framework that leverages the inferential perspective and is used for computing the result of forgetting operators is presented. It is based on the use of Ackermann's Lemma and tautology preserving formula transformations. This framework is introduced for the propositional case of forgetting and then later extended to the first-order case; 
    \item it is also shown that computing the propositional or  first-order (or fixpoint) equivalent of the dual weak forgetting operator is typically more efficient then computing the standard forgetting operator, or with computing both the weakest and strongest necessary conditions. In the light of complexity results on standard forgetting (see, e.g.,~\cite{LangLM03}), even from that one standpoint alone, it is beneficial to consider the dual weak forgetting operator as a separate operator when it is feasible to use representationally. 
\end{itemize}

The rest of the paper is structured as follows.  Section~\ref{sec:prelim} presents  preliminaries related to propositional logic and standard forgetting. Section~\ref{sec:general} discusses forgetting operators in general and provides their second-order characterization.  In Section~\ref{sec:examples}, some examples are presented illustrating the approach, where the weak forgetting operator is shown to be very useful representationally. Section~\ref{sec:sncwsc} considers the strong relationship between the dual forgetting operators and  strongest necessary and weakest sufficient conditions.  Section~\ref{sec:first-order} shows how the formalism using dual forgetting operators can be extended to the first-order case, whereas the approach to standard forgetting and its use has been predominantly propositional in nature. In Section~\ref{sec:relwork}, we discuss relevant, related work. Finally, Section~\ref{sec:concl} concludes the paper with a summary and some  final remarks.

\section{Preliminaries}\label{sec:prelim}

\subsection{Classical Propositional Logic}

For the sake of simplicity, we initially present ideas starting with classical propositional logic, \proplog, with truth constants \true (true) and \false (false), an enumerable set of propositional variables $\propvar$, and standard connectives $\lneg,\land,\lor,\implies,\equiv$. We shall also use second-order quantifiers $\exists p, \forall p$, where $p\in \propvar$. The meaning of quantifiers in the propositional context is:
   \begin{align}
      &\exists p\big(A(p)\big)\defequiv A(p=\false)\lor A(p=\true);\label{eq:exists}\\
      &\forall p\big(A(p)\big)\defequiv A(p=\false)\land A(p=\true),\label{eq:forall}
   \end{align} 
where $A(p=\expr)$ denotes a formula obtained from $A$ by substituting all occurrences of $p$ in $A$ by expression $\expr$.

By a {\em theory} we mean a finite set of formulas. A theory is identified with a conjunction of formulas it contains. We often write $\bar{p}$ to denote a tuple of propositional variables, and $Th(\bar{p})$ to indicate that theory $Th$ is formed over a vocabulary consisting of variables in $\bar{p}$. Similarly, we often write $A(\bar{p})$ to indicate that formula $A$ is formed over a vocabulary consisting of $\bar{p}$.

We say that a formula $A$ is {\em stronger (wrt $\implies$)} than a formula $B$, if $A\implies B$ is a tautology ($\models A\implies B$). In such a case, we also say that $B$ is {\em weaker (wrt $\implies$)} than $A$, $A$ is a {\em sufficient condition} for $B$, and $B$ is a {\em necessary condition} for $A$. 

A formula $A(p)$ is {\em positive} wrt $p\in\propvar$, if all occurrences of $p$ in $A$ are in the scope of an even number of negations.\footnote{As usual, we consider $B\implies C$ to stand for $\lneg B\lor C$, and $B\equiv C$ to stand for $(\lneg B\lor C)\land (B\lor\lneg C)$.} $A(p)$ is {\em negative} wrt $p\in\propvar$, if all occurrences of $p$ in $A$ are in the scope of an odd number of negations.
By a {\em literal}, we mean a propositional variable or its negation.

\subsection{Standard Forgetting}\label{sec:standard-forgetting}

Standard forgetting has been introduced in~\cite{Lin94forgetit} using a model-theoretic framework. The intuition behind this operator is to forget a part of the vocabulary of a theory and remember as much as possible using the remaining vocabulary, where logical consequences are concerned. Theorem~8 and Proposition~10 in~\cite{Lin94forgetit} provide, among others, the following important properties of forgetting, where $\forget(Th(\bar{p},\bar{q}),\bar{p})$ denotes {\em forgetting} about $\bar{p}$ in $Th(\bar{p},\bar{q})$.   

\begin{theorem}[Lin, Reiter]\label{thm:forgetit}
Let $\bar{p}$ and $\bar{q}$ be disjoint tuples of propositional variables, $A$ be a~formula not containing occurrences of variables from $\bar{p}$ and $Th(\bar{p},\bar{q})$ be a~theory. Then:
\begin{align}
   & \mbox{-- } \forget(Th(\bar{p},\bar{q}),\bar{p})\equiv \exists\bar{p}\big(Th(\bar{p},\bar{q})\big);\label{eq:forgetting-def}\\
   & \mbox{-- } Th(\bar{p},\bar{q})\models A \mbox{ iff } \forget(Th(\bar{p},\bar{q}),\bar{p})\models A.\label{eq:forgetting-thm}
\end{align}\done
\end{theorem}

Notice that~\eqref{eq:forgetting-def} can serve as an alternative definition of standard forgetting. We therefore omit discussion and use of the model-theoretic definition used in~\cite{Lin94forgetit}. 
One can also observe that due to the deduction theorem for classical propositional logic,~\eqref{eq:forgetting-thm} can be expressed as:
\begin{align}
   & \models Th(\bar{p},\bar{q})\implies A \mbox{ iff } \models \forget(Th(\bar{p},\bar{q}),\bar{p})\implies A.\label{eq:deduction}
\end{align}

\subsection{An Ackermann-Like Approach to Second-Order Quantifier Elimination}

Theorem~\ref{thm:forgetit}\eqref{eq:forgetting-def} indicates that computing forgetting using the Lin\&Reiter-like forgetting operator is equivalent to eliminating second-order quantifiers $\exists\bar{p}(\ldots)$. 
While definitions~\eqref{eq:exists}--\eqref{eq:forall} allow one to eliminate second-order quantifiers, they lead to an exponential growth of the resulting formula wrt the number of quantifiers. Therefore, as a computationally more appropriate tool, we will use the following lemma of Ackermann, already proved in~\cite{Ack} (see also~\cite{dls,gss}). This usage  typically results in much shorter formulas as a~result of quantifier eliminations in many cases.

\begin{lemma}[Propositional Ackermann Lemma]
\label{lemma:ack}
Let $A$ be a   propositional formula without occurrences of propositional variable $p$, and $B(p)$ be a propositional formula on a vocabulary containing $p$:\footnote{For the sake of clarity we assume that $B$ contains $p$, but the lemma is trivially true also when this is not the case.}
\begin{align}
& \mbox{-- if $B(p)$ is positive wrt\ $p$ then:\; }\exists p\big((p\implies
A)\land B(p)\big)\ \equiv\ B(p=A);\label{eq:ack-pos}\\
& \mbox{-- if $B(p)$ is negative wrt\ $p$ then: } \exists p\big((A\implies p)\land B(p)\big) \ \equiv\ B(p=A).\label{eq:ack-neg}
\end{align}
\done
\end{lemma}

The lemma remains true when $p$ is replaced by a second-order variable, say $P$, representing propositional formulas. For example,~\eqref{eq:ack-pos} can be formulated as:
\begin{align*}
& \mbox{-- if $B(P)$ is positive wrt\ $P$ then:\; }\exists P\big((P\implies
A)\land B(P)\big)\ \equiv\ B(P=A).
\end{align*}

Lemma~(\ref{lemma:ack}) serves as a blueprint for specifying an algorithm to eliminate 2nd-order quantifiers in many cases. It shows that if one can syntactically transform a formula $F(\bar{p})$, where $p\in \bar{p}$, into an equivalent formula with the syntactic structure on the lhs of equivalences (\ref{eq:ack-pos}) or (\ref{eq:ack-neg}), respectively, then one can eliminate $\exists p$ by substitution of $A$ for $p$ in $B$, resulting in the equivalent $B(p=A)$. This syntactic technique can be iterated for all 2nd-order quantifiers in $F(\bar{p})$, resulting in a logically equivalent propositional formula. 

To transform a formula into a form required in~\eqref{eq:ack-pos} or~\eqref{eq:ack-neg}, one can use the \dls algorithm of~\cite{dls}. For propositional formulas, one of the forms in the lefthand sides of equivalences~\eqref{eq:ack-pos} or~\eqref{eq:ack-neg} can always be obtained, thus guaranteeing removal of all 2nd-order quantifiers in an arbitrary propositional theory.

\begin{figure}[h!t]
\centering 
    \includegraphics[width=0.83\textwidth]{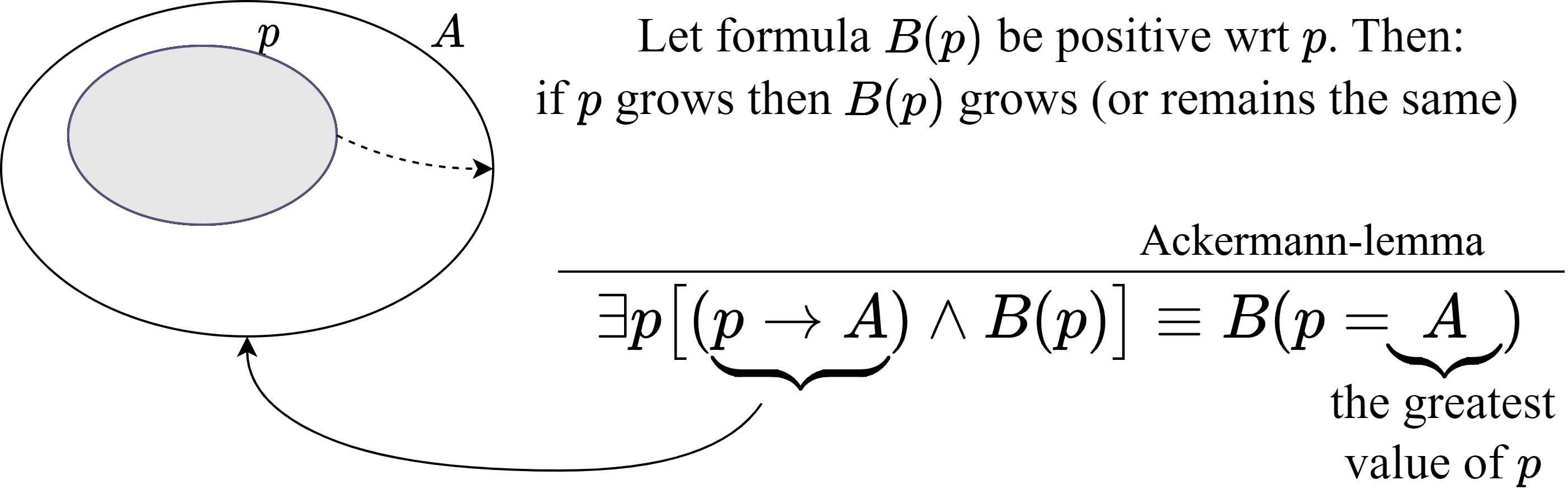}
\mbox{}\\[2em]
    \includegraphics[width=0.83\textwidth]{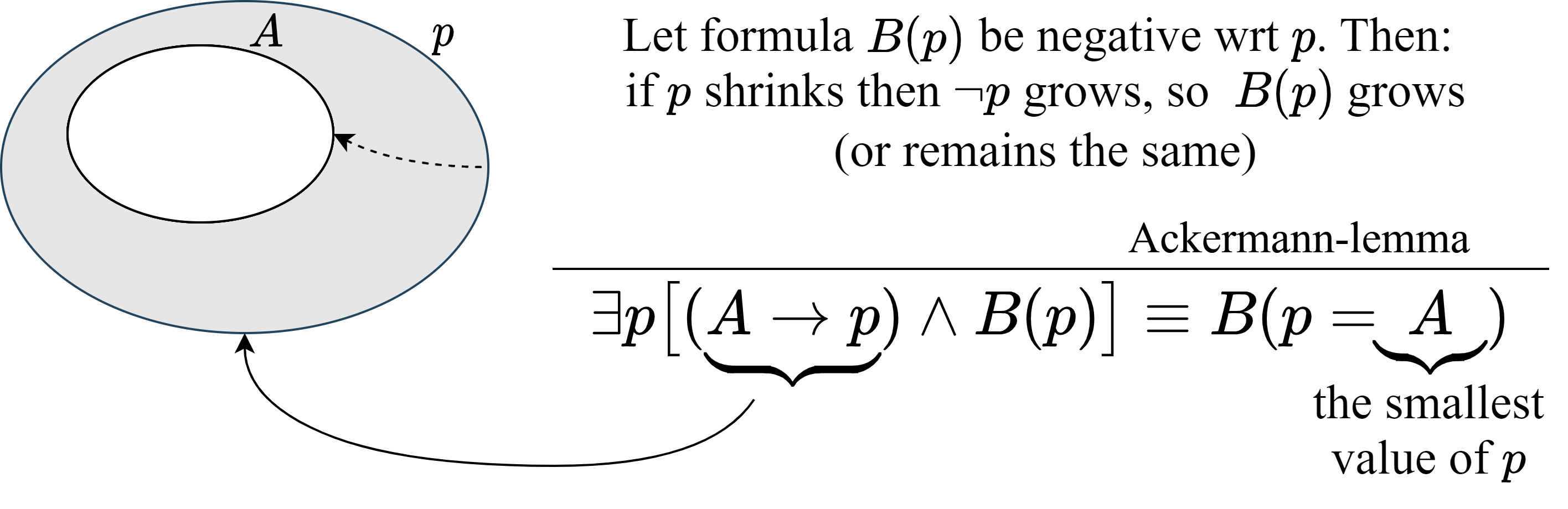}
        \caption{An illustration of intuitions behind Ackermann-like lemmas.  \label{fig:ack}}
\end{figure}

Figure~\ref{fig:ack} illustrates the idea behind Ackermann-like lemmas, where the terms ``grows'' and ``shrinks'', illustrated by dash arrows within ovals, refer to the standard ordering on truth values, $\false<\true$, compatible with the semantics of implication, which can be given by:\footnote{More formally, we deal here with the construction of a partial order among formulas for the Lindenbaum and Tarski algebra~\cite{hinman}. The Lindenbaum–Tarski algebra of a theory $T$ consists of the equivalence classes of formulas of the theory, where two formulas are equivalent when the theory $T$ proves that each implies the other. The partial order in question is then defined as:
$
||A||\leq_T ||B|| \mbox{ iff } T\vdash A\implies B,
$
where $||C||$ denotes an equivalence class of a formula $C$.} 
\[(\mbox{the truth value of } p\implies q)\defeq (\mbox{`the truth value of  $p$'}\leq \mbox{`the truth value of  $q$'}).\]

Given the constraint $p\implies A$ in~\eqref{eq:ack-pos} (or, respectively, $A\implies p$ in~\eqref{eq:ack-neg}), the greatest value of $B(p)$ is obtained when $p$ takes its greatest (respectively, smallest) value, i.e., the value given by $A$. Of course, in such cases, the existential quantifier $\exists p(\ldots)$ obtains the greatest value of $B(p)$, achieved for $p=A$. In this case, each of the constraints $p\implies A$ (in~\eqref{eq:ack-pos}) as well as $A\implies p$ (in~\eqref{eq:ack-neg}) evaluate to \true, and so disappear from the result.

\section{The Entailment-Based Inferential Perspective on Forgetting} \label{sec:general}

\subsection{Some Intuitions}

Let $Th(\bar{p},\bar{q})$ be a propositional theory over a vocabulary consisting of $\bar{p}, \bar{q}$.\footnote{When we refer to tuples of variables as arguments, like in $Th(\bar{p}, \bar{q})$, we always assume that $\bar{p}$ and $\bar{q}$ are disjoint.} When forgetting $\bar{p}$ in the theory $Th(\bar{p},\bar{q})$, one can delineate two alternative views, one existing and one new, with both resulting in a theory expressed in the vocabulary containing $\bar{q}$ only:
\begin{itemize}
    \item {\em strong (standard) forgetting} $\nforget{Th(\bar{p},\bar{q}); \bar{p}}$: a~theory that preserves the entailment of necessary conditions over $\bar{q}$;
    \item {\em weak forgetting} $\sforget{Th(\bar{p},\bar{q}); \bar{p}}$: a~theory that preserves the entailment of sufficient conditions over $\bar{q}$.
\end{itemize}   

The rationale behind the operator $\nforget{Th(\bar{p},\bar{q}); \bar{p}}$ is that one wants to remember a~theory on vocabulary $\bar{q}$, whose consequences are also consequences of the original theory. That is, for any formula $A$ on a vocabulary disjoint with $\bar{p}$,
\begin{equation}\label{eq:fnc-intuition}
    \models Th(\bar{p},\bar{q})\implies A \mbox{ iff } \models \nforget{Th(\bar{p},\bar{q});\bar{p}}\implies A.
\end{equation}
In this case, a formula $A$ is a consequence of the result of forgetting, \nforget{}, iff it is a consequence of the original theory  $Th(\bar{p},\bar{q})$. Notice the similarity of~\eqref{eq:deduction} and~\eqref{eq:fnc-intuition} indicating that $\forget()$ and $\nforget{}$ act in the same manner. However, while in the approach of~\cite{Lin94forgetit}, the property~\eqref{eq:deduction} is a~derived theorem, in our approach it is a fundamental starting point.

The rationale behind the weak forgetting operator $\sforget{Th(\bar{p},\bar{q}); \bar{p}}$ is that one wants to remember a~theory on vocabulary $\bar{q}$ such that a~formula implies $\sforget{Th(\bar{p},\bar{q}); \bar{p}}$ iff it implies the original theory. That is, for any formula $A$ on a vocabulary disjoint with $\bar{p}$,
\begin{equation}\label{eq:fsc-intuition}
\models A\implies Th(\bar{p},\bar{q}) \mbox{ iff } \models A\implies\sforget{Th(\bar{p},\bar{q});\bar{p}}.
\end{equation}
That is, a formula $A$ implies the result of weak forgetting, \sforget{}, iff it implies the original theory  $Th(\bar{p},\bar{q})$. 

\subsection{The Operator \nforget{}}\label{sec:nforget}

The requirement \eqref{eq:fnc-intuition} that \nforget{Th(\bar{p},\bar{q});\bar{p}} preserves the entailment of necessary conditions over a~vocabulary disjoint with $\bar{p}$ can be expressed as:
\begin{equation}\label{eq:nforget}
\forall P \Big(
    \forall\bar{p}\big(Th(\bar{p},\bar{q})\implies P\big) \equiv \big(\nforget{Th(\bar{p},\bar{q}); \bar{p}}\implies P\big)    
\Big),  
\end{equation}
where $P$ is a second-order variable representing an arbitrary formula over a vocabulary disjoint with $\bar{p}$. 

Using (\ref{eq:nforget}) as a basis, let's derive an important Theorem~\ref{thm:nforget}, characterizing the operator, \nforget{}. By a~standard propositional tautology, we represent equivalence $\equiv$ in~\eqref{eq:nforget}  as the conjunction of left-to-right ($\implies$) and right-to-left ($\leftarrow$) implications. 

\subsubsection{The Analysis of the Left-to-Right Direction}

Let us start with the analysis of the left-to-right direction ($\implies$) of~\eqref{eq:nforget}. Since variables $\bar{p}$ occur only in $Th(\bar{p},\bar{q})$, the implication is equivalent to:
\begin{equation}\label{eq:nforget1}
\forall P\Big(
    \big(
       \exists\bar{p}\big(Th(\bar{p},\bar{q})\big)\implies P
    \big)\implies  \big(\nforget{Th(\bar{p},\bar{q}); \bar{p}}\implies P \big)  
\Big),  
\end{equation}
In order to apply Lemma~\ref{lemma:ack}, we have to transform~\eqref{eq:nforget1} into an~equivalent form:
\begin{equation}\label{eq:nforget2}
\lneg\exists P\Big(
    \big(
       \exists\bar{p}\big(Th(\bar{p},\bar{q})\big)\implies P
    \big)\land \nforget{Th(\bar{p},\bar{q}); \bar{p}}\land\lneg P 
\Big). 
\end{equation}
We eliminate the second-order quantifier $\exists P$ from~\eqref{eq:nforget2} using Lemma~\ref{lemma:ack}\eqref{eq:ack-neg}. As a result we obtain the following formula equivalent to~\eqref{eq:nforget2}:
\begin{equation}\label{eq:nforget3}
\begin{array}{l}
\lneg\Big(
       \nforget{Th(\bar{p},\bar{q}); \bar{p}}\land\lneg\exists\bar{p}\big(Th(\bar{p},\bar{q})\big) 
\Big),  
\end{array}
\end{equation}
which in turn is equivalent to:
\begin{equation}\label{eq:nforget4}
\begin{array}{l}
       \nforget{Th(\bar{p},\bar{q}); \bar{p}} \implies\exists\bar{p}\big(Th(\bar{p},\bar{q})\big).
\end{array}
\end{equation}

\subsubsection{The Analysis of the Right-to-Left Direction}

For the right-to-left direction ($\leftarrow$) of~\eqref{eq:nforget} we proceed as follows. Since variables $\bar{p}$ occur only in $Th(\bar{p},\bar{q})$, the implication ($\leftarrow$) is equivalent to:
\begin{equation}\label{eq:nforgetr2}
	\forall P\Big(
	\big(\nforget{Th(\bar{p},\bar{q}); \bar{p}}\implies P \big)
	\implies  
	\big(
	\exists\bar{p}\big(Th(\bar{p},\bar{q})\big)\implies P
	\big)  
	\Big),  
\end{equation}
which  is equivalent to:
\begin{equation}\label{eq:nforgetr3}
	\lneg\exists  P\Big(
	\big(\nforget{Th(\bar{p},\bar{q}); \bar{p}}\implies P \big)
	\land  
\exists\bar{p}\big(Th(\bar{p},\bar{q})\big)\land\lneg  P
	\Big).
\end{equation}
As before, we eliminate the second-order quantifier $\exists P$ from~\eqref{eq:nforgetr3} using Lemma~\ref{lemma:ack}\eqref{eq:ack-neg}. As a~result, we obtain the following formula equivalent to~\eqref{eq:nforget2}:
\begin{equation}\label{eq:nforgetr4}
\lneg\Big(	\exists\bar{p}\big(Th(\bar{p},\bar{q})\big)\land \lneg \nforget{Th(\bar{p},\bar{q}); \bar{p}}
	\Big),  
\end{equation}
which in turn is equivalent to:
\begin{equation}\label{eq:nforget5}
\begin{array}{l}            \exists\bar{p}\big(Th(\bar{p},\bar{q})\big) \implies\nforget{Th(\bar{p},\bar{q}); \bar{p}}.
\end{array}
\end{equation}

\subsubsection{A Characterization of \nforget{}}
	
Using~\eqref{eq:nforget4},~\eqref{eq:nforget5} and Theorem~\ref{thm:forgetit}\,\eqref{eq:forgetting-def}, we have the following characterization of \nforget{}.

\begin{theorem}\label{thm:nforget}
For arbitrary tuples of propositional variables $\bar{p}, \bar{q}$ and $Th(\bar{p},\bar{q})$,
\begin{enumerate}
    \item 
    $\nforget{Th(\bar{p},\bar{q}); \bar{p}}  \equiv \exists \bar{p}\big(Th(\bar{p},\bar{q})\big).$
    \item $\nforget{Th(\bar{p},\bar{q}); \bar{p}} \equiv \forget(Th(\bar{p},\bar{q}); \bar{p})$.
    \item \nforget{Th(\bar{p},\bar{q}),\bar{p}} is the strongest (wrt $\implies$) formula over vocabulary $\bar{q}$, satisfying~\eqref{eq:nforget}.\done
\end{enumerate}
\end{theorem}

\subsection{The Operator \sforget{}}\label{sec:sforget}

By analogy to \nforget{}, the requirement \eqref{eq:fsc-intuition} that \sforget{Th(\bar{p},\bar{q});\bar{p}} preserves entailment by sufficient conditions can be expressed as:
\begin{equation}\label{eq:sforget}
\forall P\Big(
    \big(P\implies\sforget{Th(\bar{p},\bar{q}); \bar{p}}\big) \equiv   \forall\bar{p}\big( P\implies Th(\bar{p},\bar{q})\big)
\Big),  
\end{equation}
where $P$ is again a second-order variable representing an arbitrary formula over a vocabulary disjoint with $\bar{p}$. As in Section~\ref{sec:nforget}, we represent equivalence $\equiv$ in~\eqref{eq:sforget}  as the conjunction of two implications ($\implies$) and ($\leftarrow$).

\subsubsection{The Analysis of the Left-to-Right Direction}

Let us first consider implication ($\implies$). Since $P$ is $\bar{p}$ free, the implication is equivalent to:
\begin{equation}\label{eq:sforget1}
\forall P\Big(
    \big(P\implies\sforget{Th(\bar{p},\bar{q}); \bar{p}}\big) \implies  \big(P\implies \forall\bar{p}\big(Th(\bar{p},\bar{q})\big)\big)
\Big).  
\end{equation}
To apply Lemma~\ref{lemma:ack}, we transform~\eqref{eq:sforget1} to an equivalent form:
\begin{equation}\label{eq:sforget2}
\lneg\exists P\Big(
    \big(P\implies\sforget{Th(\bar{p},\bar{q}); \bar{p}}\big)\land   P\land\lneg \forall\bar{p}\big(Th(\bar{p},\bar{q})\big)
\Big).
\end{equation}
We eliminate the second-order quantifier  $\exists P$  from~\eqref{eq:sforget2}, using Lemma~\ref{lemma:ack}\eqref{eq:ack-pos}. As a result, we obtain the following formula equivalent to~\eqref{eq:sforget2}: 
\begin{equation}\label{eq:sforget3}
\begin{array}{l}
\lneg\Big(
   \sforget{Th(\bar{p},\bar{q}); \bar{p}}\big)\land\lneg \forall\bar{p}\big(Th(\bar{p},\bar{q})\big)
\Big),  
\end{array}
\end{equation}
which in turn is equivalent to:
\begin{equation}\label{eq:sforget4}
\begin{array}{l}
   \sforget{Th(\bar{p},\bar{q}); \bar{p}}\big)\implies \forall\bar{p}\big(Th(\bar{p},\bar{q})\big).  
\end{array}
\end{equation}

\subsubsection{The Analysis of the Right-to-Left Direction}

The implication ($\leftarrow$) of~\eqref{eq:sforget} is equivalent to:

\begin{equation}\label{eq:sforgets1}
	\forall P\Big(
	\big(P\implies \forall\bar{p}\big(Th(\bar{p},\bar{q})\big)\big)
	\implies  
	\big(P\implies\sforget{Th(\bar{p},\bar{q}); \bar{p}}\big)
	\Big).  
\end{equation}
To apply Lemma~\ref{lemma:ack}, we transform~\eqref{eq:sforgets1} to an equivalent form:
\begin{equation}\label{eq:sforgets2}
	\lneg \exists P\Big(
\big(P\implies \forall\bar{p}\big(Th(\bar{p},\bar{q})\big)\big)
\land  
P\land\lneg \sforget{Th(\bar{p},\bar{q}); \bar{p}}
\Big).
\end{equation}
We eliminate the second-order quantifier  $\exists P$ from~\eqref{eq:sforgets2}, using Lemma~\ref{lemma:ack}\eqref{eq:ack-pos}. As a result, we obtain the following formula equivalent to ~\eqref{eq:sforgets2}:
\begin{equation}\label{eq:sforgets3}
	\lneg \Big( 
\forall\bar{p}\big(Th(\bar{p},\bar{q})\big)\land\lneg \sforget{Th(\bar{p},\bar{q}); \bar{p}}
\Big). 
\end{equation}
which in turn is equivalent to:
\begin{equation}\label{eq:sforget5}
\begin{array}{l}
  \forall\bar{p}\big(Th(\bar{p},\bar{q})\big)\implies  \sforget{Th(\bar{p},\bar{q}); \bar{p}}\big).  
\end{array}
\end{equation}

\subsubsection{A Characterization of \sforget{}}

Combining \eqref{eq:sforget4} and \eqref{eq:sforget5}, we have the following characterization of \sforget{}.

\begin{theorem}\label{thm:sforget}
For arbitrary tuples of propositional variables $\bar{p}, \bar{q}$ and $Th(\bar{p},\bar{q})$,
\begin{enumerate}
    \item 
$    \sforget{Th(\bar{p},\bar{q}); \bar{p}}\equiv \forall \bar{p}\,\big(Th(\bar{p},\bar{q})\big).
$
\item \sforget{Th(\bar{p},\bar{q});\bar{p}} is the weakest (wrt $\implies$) formula over vocabulary $\bar{q}$, satisfying~\eqref{eq:sforget}.\done
\end{enumerate}
\end{theorem}

\subsection{Combining \nforget{} and \sforget{} in a dual or complementary perspective}\label{sec:best}

The net result of the analysis provided is that standard or strong forgetting, and weak forgetting are complementary in an intuitive and formally concise manner. 

Weak forgetting, \sforget{}, allows one to \textit{remember} more than using solely standard (strong) forgetting, \nforget{}, relative to a specific theory. This can be shown to be very useful in applications of forgetting operators as exhibited in Section~\ref{sec:examples}. 

In the original example~\eqref{eq:maintain}, the weak forgetting operator \sforget{} can be applied naturally,
\begin{equation}
    \sforget{(lt\lor lp);lt}\equiv \forall\,lt\,(lt\lor lp)\equiv lp
\end{equation}
As suggested in Section~\ref{sec:intro}, $lt$ is forgotten, while retaining additional information $lp$ about the physical system.

\begin{figure}[t]
\begin{centering}
    \hspace*{3.1cm}\includegraphics[scale=0.12]{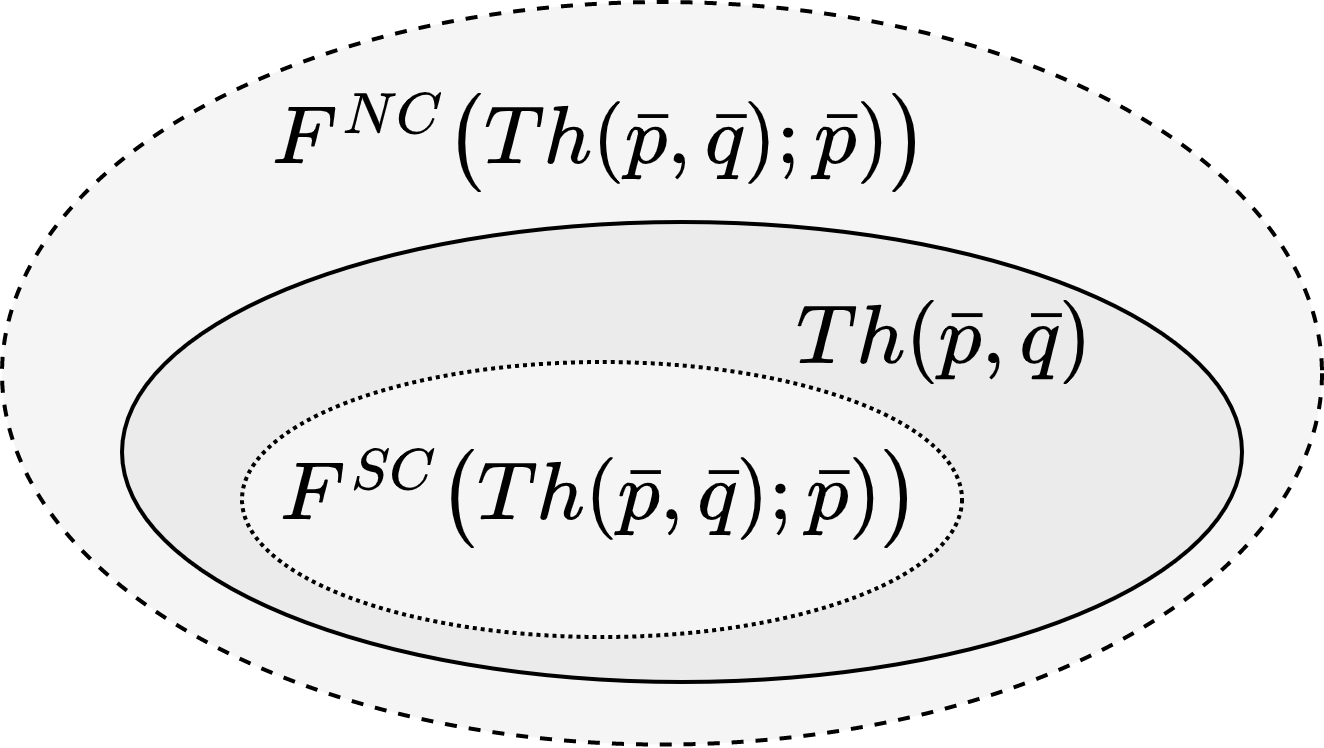}
\end{centering}
\caption{The relationships between forgetting operators and the original theory in terms of entailment: the ovals inclusion indicates that the formula in an inner oval entails the formula in an outer one.  \label{fig:approx}}
\end{figure}

It is interesting to observe that \nforget{} and \sforget{} partition the theory $Th(\bar{p},\bar{q})$ into three nicely related classes of formulas (see Figure~\ref{fig:approx}):
\begin{itemize}
    \item the innermost oval, \sforget{Th(\bar{p},\bar{q});\bar{p}}: formulas in vocabulary $\bar{q}$ that imply the original theory (sufficient conditions of the theory);
    \item the central oval: the original theory $Th(\bar{p},\bar{q})$;
    \item the outermost oval, \nforget{Th(\bar{p},\bar{q});\bar{p}}: formulas in vocabulary $\bar{q}$ implied by the original theory (necessary conditions of the theory).
\end{itemize}

From another perspective, 
\begin{itemize}
    \item $\nforget{Th(\bar{p},\bar{q}); \bar{p}}$ is ``the best'' upper approximation of $Th(\bar{p},\bar{q})$ using the restricted vocabulary (as shown in point 3 of Theorem~\ref{thm:nforget});
    \item $\sforget{Th(\bar{p},\bar{q}); \bar{p}}$ is ``the best'' lower approximation of $Th(\bar{p},\bar{q})$ using the restricted vocabulary (as shown in point 2 of Theorem~\ref{thm:sforget}).
\end{itemize}

\subsection{The Computational Perspective for  Weak Forgetting}\label{sec:overhead}

The computational overhead involved in computing weak forgetting \sforget{} is, in the worst case, the same as for computing standard forgetting \nforget{}, which is exponential in the size of the input formula, yet there are some pragmatic distinctions. Transformations of formulas made for \nforget{} can many times be reused in computing \sforget{}. Moreover, frequently there may  be a more substantial complexity gain. Notice that in some applications, theories are typically presented as sets of formulas,  $Th(\bar{p},\bar{q})=\{A_1(\bar{p},\bar{q}),\ldots, A_k(\bar{p},\bar{q})\}$,
interpreted as the conjunction
$Th(\bar{p},\bar{q})\equiv A_1(\bar{p},\bar{q})\land\ldots\land A_k(\bar{p},\bar{q}).$ 
By Theorem~\eqref{eq:deduction},

\[
\begin{array}{ll}
   \sforget{Th(\bar{p},\bar{q});\bar{p}}\equiv \forall\bar{p}\big(Th(\bar{p},\bar{q})\big)\equiv &  \forall\bar{p}\big(A_1(\bar{p},\bar{q})\land\ldots\land A_k(\bar{p},\bar{q})\big)\equiv\\
   & \forall\bar{p}\big(A_1(\bar{p},\bar{q})\big)\land\ldots\land \forall\bar{p}\big(A_k(\bar{p},\bar{q})\big).
\end{array}
\]

This partitioning allows one to eliminate second-order quantifiers $\forall{\bar{p}}$ in each conjunct separately. Though this has to be repeated $k$ times, the formulas involved are smaller, which in typical cases results in a~significant complexity reduction. 

Notice also that frequently in applications, the formulas $A_1(\bar{p},\bar{q}),\ldots$, $A_k(\bar{p},\bar{q})$ are syntactically structured in the form of rules/implications (or sequents) of the form:
\begin{equation}\label{eq:rule}
\big(\ell_1\land\ldots\land \ell_r\big)\implies\big( \ell_{r+1}\lor\ldots\lor \ell_s\big),   
\end{equation}
where $\ell_i$ ($i=1,\ldots,s$) are literals. Of course,~\eqref{eq:rule} is equivalent to $\lneg\ell_1\lor\ldots\lor \lneg\ell_r\lor \ell_{r+1}\lor\ldots\lor \ell_s.   
$
Eliminating second-order quantifiers $\bar{p}$ from:
\begin{equation}\label{eq:rule1}
\forall\bar{p}\big(\lneg\ell_1\lor\ldots\lor \lneg\ell_r\lor \ell_{r+1}\lor\ldots\lor \ell_s\big)      
\end{equation}
is then straightforward, as shown in Algorithm~\ref{alg:rule1}. 

\begin{algorithm}[ht]
\caption{Eliminating second-order quantifiers from formulas of the form~\eqref{eq:rule1}.\label{alg:rule1}}

     \KwData{Formula $A\equiv$ \eqref{eq:rule1}}
     \KwResult{Formula $B$ equivalent to $A$, without second-order quantifiers.}
     \Begin{
      $B\gets$ the formula obtained from $A$ after removing double negations\;

     \eIf{there is a literal and its negation in the resulting clause}{$B\gets \true$}{$B\gets$ the formula obtained from $B$ after removing literals involving propositional variables from $\bar{p}$, where the empty disjunction that may be obtained is \false}
         
     Remove from $B$ quantifiers binding propositional variables in $\bar{p}$ 
    }
\end{algorithm}

The following example illustrates this method.

\begin{example}
Consider the following formula:
\begin{equation}\label{eq:ex-comp}
\forall q\forall r\big(\lneg\lneg q\lor\lneg r \lor\lneg s\lor t\big).
\end{equation}
After removing double negation in front of $q$, then removing the literals involving quantified variables (i.e., $q$ and $\lneg r$), and finally removing the quantifiers, the algorithm outputs $(\lneg s\lor t)$ as a formula equivalent to~\eqref{eq:ex-comp}.
\done
\end{example}

\section{Some Examples}\label{sec:examples}

The first example shows that standard forgetting can be too strong in scenarios, where formulas (rules) depend on some  uncontrolled parameters. That is, propositions whose truth values depend on the parameter's values, occur only in the rule's premises. In such cases, when a parameter's value becomes unknown, e.g., due to a failure of a measuring device, one would like to forget the corresponding proposition as this becomes meaningless. In standard forgetting, all rules involving such propositions disappear, reducing to \true.

\begin{example}
Consider a toy expert system for maintaining a proper balance between temperature and pressure in a production process. While the pressure can be controlled, we assume that outside temperature is not controllable. Let:
\begin{itemize}
    \item $mt, ht$ stand for ``medium'' and ``high temperature'';
    \item  $lp, mp$ stand for ``maintain low'' and ``mantain medium pressure''.
\end{itemize} 
The considered theory, $Th(mt,ht,lp,mp)$, consists of the following formulas:
\begin{align}
    & mt\implies lp\lor mp;\label{eq:toy1}\\
    & ht\implies lp.\label{eq:toy2}
\end{align}
Assume the temperature sensor is broken and one wants to forget related propositions $mt, ht$. Using Theorems~\ref{thm:nforget}(1) and Theorem~\ref{thm:sforget}(1), simple transformations of formulas, and Lemma~\ref{lemma:ack}, we obtain that:
\begin{itemize}
    \item $\nforget{Th(mt,ht,lp,mp); mt,ht}$ is equivalent to \true.
    \item $\sforget{Th(mt,ht,lp,mp); mt,ht}$ is equivalent to  $(lp\lor mp)\land lp$, being
    equivalent to $lp$.
\end{itemize} 
Notice that $\sforget{}$ is much more informative: when the temperature is unknown, for safety reasons it is better to maintain low pressure ($lp$) as a sufficient condition for satisfying rules~\eqref{eq:toy1},~\eqref{eq:toy2}.
\done
\end{example}

The following example illustrates the use of forgetting when there are restrictions on accessing sensitive, personal data directly from a specific person or belief base. One gets around this by leveraging implicit information about that person through an inferential process that involves the use of forgetting.

\begin{example}\label{ex:hiding}
Assume Eve works in human resources in a company where Joe is employed. Eve faces some cultural barriers, or legislative restrictions, in asking Joe about his potential addiction to alcohol or drugs. In order to find the answer, she may use some extra knowledge/beliefs in addition to some neutral questions or observations in probing for an answer. To illustrate the approach consider a~belief, expressed by the following simple formula, where $fdd$ stands for ``frequently denies driving'', $ld$ stands for ``likes driving'', and  $pa$ stands for ``potentially addicted'':
\begin{equation}\label{eq:addict}
  fdd\implies (\lneg ld\lor pa).
\end{equation}
When using~\eqref{eq:addict} to detect Joe's potential addiction to alcohol or drugs, Eve may wonder when this formula implies $pa$, without referring to $pa$ itself. That is, she might be interested in the theory:
\[Th(pa, ld, fdd)\defequiv \big( \underbrace{(fdd \implies (\lneg ld\lor pa))}_{\mbox{\eqref{eq:addict}}}\implies pa\big),\] 
but forgetting about $pa$. In this case, weak forgetting can be used:
\begin{equation}\label{eq:pa}
    \sforget{Th(pa, ld, fdd);pa}.
\end{equation}
According to Theorem~\ref{thm:sforget}(1), \eqref{eq:pa} is equivalent to:
\begin{equation}
    \forall pa\big((fdd\implies  (\lneg ld\lor pa))\implies pa\big),
\end{equation}
which is equivalent to:
\begin{equation}\label{eq:paexists}
    \lneg \exists pa\big((fdd\implies  (\lneg ld\lor pa))\land \lneg pa\big).
\end{equation}
Formula~\eqref{eq:paexists} can easily be transformed into a form required for application of Lemma~\ref{lemma:ack}:
\[
\lneg \exists pa\big((\underbrace{pa\implies\false}_{\lneg pa})\land (fdd\implies  (\lneg ld\lor pa))\big).
\]
Now Lemma~\ref{lemma:ack}\eqref{eq:ack-pos} results in the following equivalent formula:
\begin{equation}
    \lneg \big(fdd\implies  (\lneg ld\lor \false)\big)\equiv
     \lneg \big(fdd\implies  \lneg ld\big)\equiv
     fdd\land ld.
\end{equation}
To derive the conclusion that Joe is potentially addicted ($pa$), Eve might try to find out independently, whether Joe frequently denies driving while at the same time likes driving. If both these assertions are true, then $pa$ is true too.

Observe that \nforget{Th(pa, ld, fdd);pa} is equivalent to:
\[
\exists pa\big((fdd\implies  (\lneg ld\lor pa))\implies pa\big),
\]
which is equivalent to \true (with witness $pa\defequiv\true$). So, it could not be used in this particular context.\done
\end{example}

The last example in this section is inspired by applying forgetting to the problem of restoring consistency of belief bases merged from different sources~\cite{LangM10}.

\begin{example}\label{ex:tennisswimming}
	Assume that Jack (the first source) considers building an outdoor recreational complex consisting of tennis courts ($tc$) and a~swimming pool ($sp$). He allocates a~budget for the investments. Let `$bdg$' expresses that the overall investment fits in the budget, `$loan$' stand for getting a loan and `$inv$' denote involvement of an additional investor. Jack's  requirements are then expressed as: 
	\begin{equation}\label{eq:tennisswimming}
		tc\land sp\land (bdg\lor loan\lor inv).
	\end{equation} 
He hires an external consultant (the second source)  who believes that in addition to outdoor sport facilities, an indoor squash court ($isq$) and a gym center ($gc$) are necessary to attract enough customers. This can be expressed by $\!\!\!\!\!\!\!\!\!\underbrace{(tc\lor sp)}_{\mbox{outdoor facilities}}\!\!\!\!\!\!\!\!\!\implies (isq\land gc)$. The consultant is well aware of specific costs for the different centers and courts.
Consequently, the consultant believes that building an additional indoor squash court will not require a~loan ($isq\implies\lneg loan$), while for a gym center a~loan will be needed ($gc\implies loan$). Summing up, the external consultant's beliefs can be expressed as:
\begin{equation}\label{eq:consultant}
	\big((tc\lor sp)\implies (isq\land gc)\big) \land \big(isq\implies \lneg loan\big)\land \big(gc\implies loan\big).
\end{equation}
Jack's beliefs, expressed by~\eqref{eq:tennisswimming}, merged with the consultant's beliefs, which are expressed by~\eqref{eq:consultant}, are jointly inconsistent on  `$loan$'. The investor decides to forget about `$loan$'. According to consistency restoring strategies considered in~\cite{LangM10}, in order to derive meaningful conclusions, Jack can forget about `$loan$' in selected parts of the merged belief bases. Assume that he considers forgetting `$loan$' in~\eqref{eq:tennisswimming} or~\eqref{eq:consultant}:
\begin{align}
	\mbox{-- } & \nforget{\{\eqref{eq:tennisswimming}\}; loan}\equiv tc\land sp,\label{eq:ex3an}\\  
	 &\qquad \sforget{\{\eqref{eq:tennisswimming}\}; loan}\equiv tc\land sp\land (bdg\lor inv);\label{eq:ex3as}\\ 
     \mbox{-- } & \nforget{\{\eqref{eq:consultant}\}; loan}\equiv 
     \big((tc\lor sp)\implies (isq\land gc)\big) \land \big(isq\implies  \lneg gc), \label{eq:ex3bn} \\
	 &\qquad  \sforget{\{\eqref{eq:consultant}\}; loan}\equiv \big((tc\lor sp)\implies (isq\land gc)\big) \land  \lneg isq\land \lneg gc.\label{eq:ex3bs}
\end{align}

Notice that weak forgetting \sforget{} adds  substantial additional informative content in comparison to standard forgetting  $\nforget{}$, by providing sufficient conditions for the respective formulas:
\begin{itemize}
	\item \eqref{eq:ex3as} extends \eqref{eq:ex3an} by information that to satisfy the formula~\eqref{eq:tennisswimming} after forgetting about `$loan$', it suffices that additionally $(bdg\lor inv)$ holds;
	\item \eqref{eq:ex3bs} extends \eqref{eq:ex3bn} by information that to satisfy the formula~\eqref{eq:consultant} after forgetting about `$loan$', it suffices that additionally  $\lneg isq$ and $\lneg gc$ hold.
 \done
\end{itemize} 
\end{example}

Although the examples considered may seem somewhat contrived due to their brevity, they do target deeper generic applications in which forgetting operators can play a substantial inferential role. Additionally, the pragmatic use and importance of the weak forgetting operator is clearly shown as a useful inferential complement to the standard forgetting operator. 

\section{Relationship with Strongest Necessary and Weakest Sufficient Conditions}\label{sec:sncwsc}

Strongest necessary and weakest sufficient conditions have been introduced in~\cite{Lin}. Let us recall the definitions with minor adjustments.

By {\em a necessary condition
of a formula $ A(\bar{p},\bar{q})$ on propositional variables $\bar{q}$ under theory $Th(\bar{p},\bar{q})$}
we shall understand any formula
$B(\bar{q})$ containing only symbols in $\bar{q}$ such that\break 
$T\models  A(\bar{p},\bar{q})\implies B(\bar{q})$. Such a formula $B(\bar{q})$ is the {\em strongest necessary condition}, denoted by $\snc{Th(\bar{p},\bar{q});  A(\bar{p},\bar{q});\bar{q}}$ if, additionally,
for any necessary condition $C(\bar{q})$ of $ A(\bar{p},\bar{q})$ on $\bar{q}$ under $Th(\bar{p},\bar{q})$, we
have that $T(\bar{p},\bar{q})\models  B(\bar{q})\implies C(\bar{q})$.

By {\em a sufficient condition
of a formula $ A(\bar{p},\bar{q})$ on propositional variables $\bar{q}$ under theory $Th(\bar{p},\bar{q})$}
we shall understand any formula
$B(\bar{q})$ containing only symbols in $\bar{q}$ such that $Th(\bar{p},\bar{q})\models B(\bar{q})\implies  A(\bar{p},\bar{q})$. It is the {\em weakest sufficient condition},
denoted by $\wsc{Th(\bar{p},\bar{q}); A(\bar{p},\bar{q});\bar{q}}$ if, additionally, for any sufficient condition $C(\bar{q})$
of $A(\bar{p},\bar{q})$ on $\bar{q}$ under $Th(\bar{p},\bar{q})$, we
have that $Th(\bar{p},\bar{q})\models C(\bar{q})\implies B(\bar{q})$.

The following second-order characterization of \snc{} and \wsc{} has been provided in~\cite{dlssnc}.

\begin{lemma}\label{lemma:wsc-snc-so}
For any $ A(\bar{p},\bar{q})$, $\bar{q}$ and $T(\bar{p},\bar{q})$:
\begin{align}
&\;\snc{Th(\bar{p},\bar{q}); A(\bar{p},\bar{q});\bar{q}}\equiv \exists \bar{p}\big(Th(\bar{p},\bar{q})\land  A(\bar{p},\bar{q})\big);\label{eq:so-snc}\\
&\;\wsc{Th(\bar{p},\bar{q}); A(\bar{p},\bar{q});\bar{q}} \equiv \forall \bar{p}\big(Th(\bar{p},\bar{q})\implies  A(\bar{p},\bar{q})\big). \label{eq:so-wsc}
\end{align}
\done
\end{lemma}

By use of Theorem~\ref{thm:nforget}(1), Theorem~\ref{thm:sforget}(1) and Lemma~\ref{lemma:wsc-snc-so}, the following corollary shows that the dual forgetting operators and strongest necessary and weakest sufficient conditions are mutually definable.

\begin{corollary}\label{prop:forget-snc-wsc}\mbox{}
\begin{align}
  &\;\snc{Th(\bar{p},\bar{q}); A(\bar{p},\bar{q});\bar{q}}\!\equiv\!
  \nforget{Th(\bar{p},\bar{q})\land  A(\bar{p},\bar{q});\bar{p}};\label{eq:corollarysnc}\\
  &\;\wsc{Th(\bar{p},\bar{q}); A(\bar{p},\bar{q});\bar{q}}\!\equiv\!
\sforget{Th(\bar{p},\bar{q})\!\implies\!A(\bar{p},\bar{q});\bar{p}};\label{eq:corollarywsc}\\
  &\; \nforget{Th(\bar{p},\bar{q});\bar{p}}\equiv\snc{Th(\bar{p},\bar{q});\true;\bar{q}}; \\  
  &\; \sforget{Th(\bar{p},\bar{q});\bar{p}}\equiv\wsc{\lneg Th(\bar{p},\bar{q});\false;\bar{q}}.
\end{align} 
\done
\end{corollary}

Even though Corollary~\ref{prop:forget-snc-wsc} exists, notice that forgetting operators typically serve  purposes different than those of strongest necessary and weakest sufficient conditions. That is, the operators \snc{} and \wsc{} focus on queries $A(\bar{p},\bar{q})$ to,  or observations complementing a theory $Th()$,  when the underlying language is reduced. In applying \nforget{} and \sforget{}, one looks for the ``best'' approximations of the theory in a reduced sublanguage. For example, 
\begin{itemize}
    \item the weakest sufficient condition \wsc{Th(\bar{p},\bar{q}); A(\bar{p},\bar{q});\bar{q}} can be used to \textit{explain  the query (observation)}, $A(\bar{p},\bar{q})$,  given the theory $Th(\bar{p},\bar{q})$, in a sublanguage consisting of $\bar{q}$;
    \item on the other hand, the associated weak forgetting operator \sforget{Th(\bar{p},\bar{q});\bar{p}},
    can be used to \textit{explain the theory} $Th(\bar{p},\bar{q})$
    in the same sublanguage, consisting of $\bar{q}$.
\end{itemize}

 As an added distinction, computing \sforget{Th(\bar{p},\bar{q});\bar{p}} is usually more efficient than computing \wsc{Th(\bar{p},\bar{q}); A(\bar{p},\bar{q});\bar{q}} (see also Sections~\ref{sec:overhead} and~\ref{sec:compasp}).

\section{A First-Order Extension} \label{sec:first-order}

Generalization of forgetting operators to the first-order case is an important topic of research and also essential for many applications~\cite{Delgrande17,Eiter-kern-isberner,Lin94forgetit}.  Although limited work has been done with standard forgetting, there are no general, generic approaches in this respect. In approaching this topic, classical  first- and second-order logics~\cite{vandalen,gss} will be used, assuming the following languages:
\begin{itemize}
\item classical first-order logic, $\folog$, extending propositional logic \proplog  with the set \fovar of {\em individual variables} representing domain objects, \calr of relation symbols, and first-order quantifiers $\forall, \exists$ binding individual variables;
\item second-order logic, \solog, extending \folog with {\em relational variables}, \sovar, and second-order quantifiers, also denoted by $\forall, \exists$, but binding relational variables.
\end{itemize}

Some additional terminology is also required.
The part of a logical formula to which a~quantifier is applied is called the {\em scope} of this quantifier. A quantifier {\em binds} its variable within its scope. An occurrence of a variable in a formula is bound if it occurs in the scope of a quantifier binding the same variable. Otherwise the occurrence of this variable is {\em free}. A formula not containing free variables is called {\em closed}.

\subsection{First-Order Standard and Weak Forgetting}

When shifting from entailment \eqref{eq:forgetting-thm} to implication \eqref{eq:deduction}, the deduction theorem for propositional logic was applied. In first-order logic, the deduction theorem requires that the formulas moved from the left-hand side to the right-hand side of $\models$, are closed. That is, in the first-order case there will be the  requirement that the theories considered contain only closed formulas. In practice, this is not really a restriction. Typically, belief bases are defined using closed formulas. Even if the formulas in question contain free variables, such as in rule-based theories, they are assumed to be implicitly universally quantified.

Given the above requirement pertaining to closed  theories, considerations about standard and weak forgetting provided in Section~\ref{sec:general}, including Theorems~\ref{thm:nforget} and~\ref{thm:sforget}, are preserved. This is formulated in the following theorems, where for tuples of relation symbols $\bar{r}, \bar{s}$, $Th(\bar{r},\bar{s})$ is a~closed first-order theory over vocabulary $\bar{r}, \bar{s}$ and, rather than in the propositional version~\eqref{eq:nforget}, \eqref{eq:sforget}, there is the requirement that for every second-order variable $P$ representing first-order formulas over a vocabulary disjoint with~$\bar{r}$:
\begin{itemize}
    \item \nforget{Th(\bar{r},\bar{s});\bar{r}} preserves the entailment of necessary conditions:
\begin{equation}\label{eq:nforget-fo}
\forall P \Big(
    \forall\bar{r}\big(Th(\bar{r},\bar{s})\implies P \big) \equiv \big(\nforget{Th(\bar{r},\bar{s}); \bar{r}}\implies P\big)    
\Big);  
\end{equation}
    \item \sforget{Th(\bar{p},\bar{q});\bar{p}} preserves the entailment by sufficient conditions:
\begin{equation}\label{eq:sforget-fo}
\forall P\Big(
    \big(P\implies\sforget{Th(\bar{r},\bar{s}); \bar{r}}\big) \equiv   \forall\bar{r}\big( P\implies Th(\bar{r},\bar{s})\big)
\Big).
\end{equation}
\end{itemize}

\begin{theorem}\label{thm:nforget-fo}
For arbitrary tuples of relation symbols $\bar{r},\bar{s}$ and a closed first-order theory $Th(\bar{r},\bar{s})$,
\begin{enumerate}
    \item 
    $\nforget{Th(\bar{r},\bar{s}); \bar{r}}  \equiv \exists \bar{r}\,\big(Th(\bar{r},\bar{s})\big).$
    \item $\nforget{Th(\bar{r},\bar{s}); \bar{r}} \equiv \forget(Th(\bar{r},\bar{s}); \bar{r})$.
    \item \nforget{Th(\bar{r},\bar{s}),\bar{r}} is the strongest (wrt $\implies$) formula over vocabulary $\bar{s}$, satisfying~\eqref{eq:nforget-fo}.\done
\end{enumerate}
\end{theorem}

Notice that similarly to the the case of Theorem~\ref{thm:nforget}.2, Theorem~\ref{thm:nforget-fo}.2 follows from the first statement in the above theorem together with Theorem~8 of~\cite{Lin94forgetit}.

\begin{theorem}\label{thm:sforget-fo}
For arbitrary tuples of relation symbols $\bar{r},\bar{s}$ and a closed first-order theory $Th(\bar{r},\bar{s})$,
\begin{enumerate}
    \item \mbox{}
$    \sforget{Th(\bar{r},\bar{s}); \bar{r}}\equiv \forall \bar{r}\big(Th(\bar{r},\bar{s})\big).
$
\item \sforget{Th(\bar{r},\bar{s});\bar{r}} is the weakest (wrt $\implies$) formula over vocabulary $\bar{s}$, satisfying~\eqref{eq:sforget-fo}.\done
\end{enumerate}
\end{theorem}

\subsection{First-Order Ackermann Lemma}

As in the propositional case, the first-order Ackermann Lemma provides a~powerful technique for eliminating second-order quantifiers from standard and weak forgetting formulas. The \dls algorithm of~\cite{dls} uses Lemma~\ref{lemma:ack-fo}, described below.

In order to formulate the first-order version of the Ackermann lemma, let us extend the notation $A(p=\expr)$ form Section~\ref{sec:prelim}. If $A$ is a formula, $r$ is a $k$-argument relation symbol occurring in $A$, $\expr(x_1,\ldots,x_k)$ is an expression including variables among $x_1,\ldots,x_k$, then:
\[A\big(r(x_1,\ldots,x_k)=\expr(x_1,\ldots,x_k)\big)\]  denotes a formula obtained from $A$ by substituting all occurrences of $r$ by $\expr(x_1,\ldots,x_k)$ in which variables $x_1,\ldots,x_k$ are in each instance replaced by arguments of the occurrence of $r$ being substituted. For example, let $A\!\defeq\! \big(s(x_1,a)\lor r(a,b)\lor r(b,c)\big)$ and $\expr(x_1,x_2)\!\defeq\! s(x_1,x_2)\land t(x_2,d)$. When $r(x_1,x_2)=\expr(x_1,x_2)$, we have: 
\[
`r(a,b)=\expr(a,b)\mbox{'\;is\;} s(a,b)\land t(b,d) \mbox{\;\; and \;\;}`r(b,c)=\expr(b,c)\mbox{'\;is\;}s(b,c)\land t(c,d).
\]  
Therefore we have:
\[
A\big(r(x_1,x_2)=\underbrace{(s(x_1,x_2)\land t(x_2,d))}_{\expr(x_1,x_2)}\big)=s(x_1,a)\lor \underbrace{(s(a,b)\land t(b,d))}_{r(a,b)=\expr(a,b)}\lor \underbrace{(s(b,c)\land t(c,d))}_{r(b,c)=\expr(b,c)}
\]

We can now formulate the first-order version of the Ackermann lemma.

\begin{lemma}[First-order Ackermann Lemma]\label{lemma:ack-fo}
Let $r$ be a $k$-ary relation symbol, $A$ be a  first-order formula without occurrences of $r$, $B$ be a first-order formula and $\bar{x}$ be a $k$-tuple of distinct variables. Then:
\begin{align}
& \mbox{-- if $B$ is positive wrt\ $r$ then:\; }\exists r\big(\forall\bar{x}(r(\bar{x})\implies
A(\bar{x}))\land B\big)\ \equiv\ B(r(\bar{x})=A(\bar{x}));\label{eq:ack-pos-fo} \\
& \mbox{-- if $B$ is negative wrt\ $r$ then: } \exists r\big(\forall\bar{x}(A(\bar{x})\implies r(\bar{x}))\land B\big) \ \equiv\ B(r(\bar{x})=A(\bar{x})).\label{eq:ack-neg-fo}
\end{align}\done
\end{lemma}

Observe that Figure~\ref{fig:ack} also illustrates Lemma~\ref{lemma:ack-fo}, where $A$ grows when $r$ grows (respectively shrinks).\footnote{In Figure~\ref{fig:ack}, $r$ is represented by $p$.} In the first-order case, Figure~\ref{fig:ack} is even more intuitive since growing and shrinking pertains to the tuples that satisfy the relation $r$ ($p$ in the figure), which are maximized (minimized) relative to~\eqref{eq:ack-pos-fo} and~\eqref{eq:ack-neg-fo}, respectively.

To transform a formula into a form required in~\eqref{eq:ack-pos-fo} or~\eqref{eq:ack-neg-fo}, one can again use the \dls algorithm~\cite{dls}. Unlike the propositional case, in the first-order case such a transformation is not always doable in general, but it has been shown to work for large classes of formulas~\cite{Conradie06,dls,gss}.

The following example illustrates the use of Lemma~\ref{lemma:ack-fo} in the context of standard and weak forgetting.

\begin{example}\label{ex:test}
To illustrate the use of the first-order Ackermann lemma, consider the following belief base, where $ms(x)$ stands for ``person $x$ has mild symptoms of a disease'', $ss(x)$  -- for ``person $x$ has severe symptoms of the disease'', $h(x)$ -- for ``$x$ should stay home'',  $t(x)$ -- for \mbox{``$x$ needs a test for the disease''}, and $ich(x)$ -- for ``$x$ should immediately consult a health care provider'':
\begin{equation}\label{eq:test}
    Th(ms, h, t, ss, ich)=\Big\{\forall x \Big(ms(x)\implies \big(h(x)\land t(x)\big) \Big),\;
    \forall x\Big(\big(ss(x)\lor t(x)\big)\implies ich(x)\big) \Big)\Big\}.
\end{equation}
When a test is not available, it is useful to forget about it, so one can consider standard and weak forgetting operators \nforget{Th\big(ms, h, t, ss, ich\big); t)} and \sforget{Th\big(ms, h, t, ss, ich\big); t)}:
\begin{itemize}
    \item \mbox{}\\[-3.5em] 
\begin{equation}\label{eq:testacknc}
\!\!\!\!\!\!\!\!\!\!\!\!\!\!\!\!\!\!
\begin{array}{ll}
    \nforget{Th\big(ms, h, t, ss, ich\big); t)}\equiv  \exists t\Big(&\!\!\!\!\!\!\forall x \Big(ms(x)\implies \big(h(x)\land t(x)\big) \Big)\land \\
    & \!\!\!\!\!\!\forall x\Big(\big(ss(x)\lor t(x)\big)\implies ich(x)\Big)\Big).
\end{array}
\end{equation}
In order to eliminate $\exists t$ from~\eqref{eq:testacknc} using Lemma~\ref{lemma:ack-fo}, it suffices to transform it to the equivalent form:\footnote{Notice that in this case, the form can be automatically obtained using the \dls algorithm.}
\begin{equation}\label{eq:testacknc1}
  \exists t\Big(\forall x\Big(ms(x)\implies t(x)\Big) \land\forall x \Big(ms(x)\implies h(x)\Big)\land 
\forall x\Big(\big(ss(x)\lor t(x)\big)\implies ich(x)\Big)
\Big).
\end{equation}
An application of Lemma~\ref{lemma:ack-fo}\eqref{eq:ack-neg-fo} results in:
\begin{equation}\label{eq:testacknc2}
  \forall x \Big(ms(x)\implies h(x)\Big)\land 
\forall x\Big(\big(ss(x)\lor ms(x)\big)\implies ich(x) \Big).
\end{equation}

That is, a person with minor symptoms should stay at home. If, under the circumstances, it is not possible to test for the disease, then the severe \textit{or} mild symptoms should suffice to immediately consult (by phone or a visit) the health care provider.  
Intuitively, both causes are correct:
\begin{itemize}
\item by the second formula of the belief base~\eqref{eq:test}, severe symptoms suffice for a need to immediately consult a health care provider;
    \item by the first formula of the belief base~\eqref{eq:test}, mild symptoms imply a need for making a~test which, in turn, by the second formula of the belief base, suffices for a need to immediately consult a health care provider. 
\end{itemize}
    \item \mbox{}\\[-3.5em]
\begin{equation}\label{eq:testacksc}
\!\!\!\!\!\!\!\!\!\!\!\!\!\!\!\!\!\!    
\begin{array}{ll}
    \sforget{Th\big(ms, h, t, ss, ich\big); t)}\equiv  \forall t\Big(&\!\!\!\!\!\!\forall x \Big(ms(x)\implies \big(h(x)\land t(x)\big) \Big)\land \\
    & \!\!\!\!\!\!\forall x\Big(\big(ss(x)\lor t(x)\big)\implies ich(x)\Big)\Big).
\end{array}
\end{equation}
After a transformation similar to~\eqref{eq:testacknc1}, and distributing $\forall t$ over conjunctions, one obtains the following equivalent formula:
\begin{equation}\label{eq:testacksc1}
  \forall t\forall x\Big(ms(x)\implies t(x)\Big) \land\forall t\forall x \Big(ms(x)\implies h(x)\Big)\land 
\forall t\forall x\Big(\big(ss(x)\lor t(x)\big)\implies ich(x) \Big).
\end{equation}
One can now eliminate each conjunct separately:
\begin{itemize}
    \item $\forall t\forall x\Big(ms(x)\implies t(x)\Big)\!\equiv\!\forall x\lneg\exists t\Big(ms(x)\land\lneg t(x)\Big)\!\equiv\! \forall x\lneg\exists t\Big(\forall x\big(\false\implies t(x)\big)\land ms(x)\land\lneg t(x)\Big)$.

Notice that subformula $\forall x\big(\false\implies t(x)\big)$ is added artificially to make the Ackermann Lemma work. It is actually a tautology so, in conjunction with the other formulas, it does not affect their truth value. Now an application of Lemma~\ref{lemma:ack-fo}\eqref{eq:ack-neg-fo}, results in
$\forall x\lneg\Big(ms(x)\land\lneg \false\Big)$, which is equivalent to $\forall x\Big(\lneg ms(x)\Big)$;
\item in $\forall t\forall x \Big(ms(x)\implies h(x)\Big)$ the quantifier $\forall t$ is redundant, so can simply be removed;
\item $\forall t\forall x\Big(\big(ss(x)\lor t(x)\big)\implies ich(x) \Big)\equiv \forall x\lneg\exists t\Big(\big(ss(x)\lor t(x)\big)\land\lneg ich(x) \Big)$. We add an artificial conjunct $\forall x\Big(t(x)\implies\true\Big)$ which is a tautology, apply Lemma~\ref{lemma:ack-fo}\eqref{eq:ack-pos-fo}, and obtain an equivalent formula $\forall x\lneg\Big(\big(ss(x)\lor \true\big)\land\lneg ich(x) \Big)$. This is equivalent to $\forall x\Big(ich(x)\Big)$.
\end{itemize}
Thus, we obtain the following first-order result, equivalent to~\eqref{eq:testacksc1}: 
\begin{equation}\label{eq:testnotms}
\forall x\Big(\lneg ms(x)\Big)\land \forall x\Big( ms(x)\implies h(x)\Big)\land \forall x\Big(ich(x)\Big),
\end{equation}
which is equivalent to $\forall x\Big(\lneg ms(x)\Big)\land \forall x\Big(ich(x)\Big)$. Indeed, without access to a test, $t(x)$, one can only make sure that the first formula of~\eqref{eq:test} is true by assuring  that $\forall x\Big(\lneg ms(x)\Big)$ is true. The second formula of~\eqref{eq:test} can only be guaranteed when $\forall x\Big(ich(x)\Big)$ is true. Though  $\forall x\Big(\lneg ms(x)\Big)\land \forall x\Big(ich(x)\Big)$ looks rather useless, it can actually be used to select persons satisfying $\lneg ms(x)\land ich(x)$, thus also satisfying~\eqref{eq:test}. That is, when one reduces the domain to objects satisfying $\big(\lneg ms(x)\land ich(x)\big)$, the initial theory is guaranteed to hold for those individuals.  
 \done
\end{itemize}
\end{example}

The use of \sforget{} in Example~\ref{ex:test} has interesting generic potential in that it allows one to isolate a subdomain of individuals ensuring the validity of a restricted part of the original theory in question.

\subsection{The Fixpoint Lemma}

Notice that Lemma~\ref{lemma:ack-fo} requires that formula $A$ does not contain occurrences of the eliminated relation symbol. On the other hand, when it does, one may still obtain useful results for a large class of formulas. To see this, consider once again the cases shown in Figure~\ref{fig:ack}, where rather than $A$ we consider $A(p)$. As is well known,\footnote{See, e.g.,~\cite{AHV}.} given that $A(p)$ is positive wrt $p$, there exists a~smallest and  greatest fixpoint for $A(p)$ wrt $p$. This observation is all that is required to generalize the first-order extension. The idea is formulated in the following  lemma, proved in~\cite{NS} (see also, e.g.,~\cite{dlssnc,gss}), where:
\begin{itemize}
    \item \lfp{p}{A(p)} is the \textit{least} wrt implication (that is, in the terminology used in the paper, the \textit{strongest}) formula $B$ being a fixpoint of $A(p)$, i.e., satisfying $\models B\equiv A(B)$;
    \item \gfp{p}{A(p)} is the \textit{greatest} wrt implication  (that is, in the terminology used in the paper, the \textit{weakest}) formula being a fixpoint of $A(p)$.
\end{itemize}

Intuitively, when the extension of a formula \textit{shrinks}, it becomes \textit{stronger} because it implies more formulas. Likewise, when the extension of a formula \textit{grows}, it becomes \textit{weaker} because it is implied by more formulas. Therefore, \textit{least} wrt to implication means \textit{strongest} and \textit{greatest} wrt implication means \textit{weakest}.

\begin{lemma}[Fixpoint Lemma]\label{lemma:fixpoint}
Let $r$ be a $k$-ary relation symbol, $A(r)$ be a  first-order formula with positive occurrences of $r$ only, $B$ be a first-order formula, and $\bar{x}$ be a $k$-tuple of distinct variables. Then:
\begin{align}
& \mbox{-- if $B$ is positive wrt\ $r$ then:\; }\exists r\big(\forall\bar{x}(r(\bar{x})\implies
A(r))\land B\big)\ \equiv\ B\big(r(\bar{x})=\gfp{r(\bar{x})}{A(r)}\big);\label{eq:ack-pos-fix} \\
& \mbox{-- if $B$ is negative wrt\ $r$ then: } \exists r\big(\forall\bar{x}(A(r)\implies r(\bar{x}))\land B\big) \ \equiv\  B\big(r(\bar{x})=\lfp{r(\bar{x})}{A(r)}\big).\label{eq:ack-neg-fix}
\end{align}\done
\end{lemma}
Notice that Figure~\ref{fig:ack} applies also to the fixpoint case formulated in Lemma~\ref{lemma:fixpoint}, where $A$ grows when $r$ grows (respectively shrinks).\footnote{As in the first-order case, in Figure~\ref{fig:ack} $r$ is represented by $p$.} Therefore we look for the greatest (respectively least) $r$ such that $r\equiv A(r)$, i.e., $r$ is a (greatest or least) fixpoint of $A(r)$.

The following example illustrates an application of this lemma.

\begin{example} A communication network is being designed. Due to a specific application area, the designers have to consider special security requirements. In particular they consider two networks: an internal and an external one. The internal network nodes should not be externally reachable unless they are protected by a specialized expensive security component. A part of the underlying belief base contains, among others,	the following formula, where  $con(x,y)$ stands for ``nodes $x$ and $y$ are directly connected'', and $r(x,y)$ stands for ``$y$ is reachable from $x$'':\footnote{Notice that formula \eqref{eq:seceample} can be obtained as a translation of a rule defining $r()$ as a transitive closure of $con()$.}
    \begin{equation}\label{eq:seceample}
		\forall x \forall y\Big(\big(con(x,y)\lor\exists z \big(con(x,z)\land 
			r(z,y)\big)\big)\implies r(x,y)\Big),
	\end{equation}
	 Due to the security requirements, the following integrity constraint has to be preserved when choosing direct network connections, where $ex(x)$ and $in(x)$ denote that node $x$ belongs to the external or internal network, respectively, and  $sec(x)$ denotes that node $x$ is equipped with the security component:
	 \begin{equation}\label{eq:seceamplenconstr}
	  \forall y	\Big(\exists x\big(ex(x)\land r(x,y)\big)\implies \big(in(y)\implies sec(y) \big)\Big).
	 \end{equation}
 That is, when $y$ is reachable from a node $x$ of an external network ($\exists x\big(ex(x)\land r(x,y)\big)$) then whenever $y$ is an internal node ($in(y)$) then it is to be equipped with the security protecting component ($sec(y)$). 
To focus on the design of $con()$, the designers prefer to abstract from $r()$ for the time being, so forget about $r$. 
	
Let us first compute \nforget{Th(con,r,ex,in,sec);r}, where $Th(con,r,ex,in,sec)$ denotes the conjunction $\mbox{\eqref{eq:seceample}}\land\mbox{\eqref{eq:seceamplenconstr}}$:
\begin{align}
\nforget{Th(con,r,ex,\,&in,sec);r}\equiv\nonumber\\  
    & \exists r\, \Big(\forall x \forall y\Big(\big(con(x,y)\lor\exists z \big(con(x,z)\land 
		r(z,y)\big)\big)\implies r(x,y)\Big)\land\label{eq:sec1}\\
		&\;\;\;\;\;\;\forall y	\Big(\exists x\big(ex(x)\land r(x,y)\big)\implies \big(in(y)\implies sec(y) \big)\Big).\label{eq:sec2}
	\end{align}
Observe that Lemma~\ref{lemma:ack-fo} cannot be applied due to~\eqref{eq:sec1}. However, Lemma~\ref{lemma:fixpoint}\eqref{eq:ack-neg-fix} can still be used and results in:
\begin{equation}\label{eq:lfpexample}
    \begin{array}{ll}
		\forall y	\Big(\exists x\big(ex(x)\land \lfp{r(x,y)}{con(x,y)\lor  \exists z \big(con(x,z)\,\land &\!\!\!\!
			r(z,y)\big)}  \big)\\& \implies \big(in(y)\implies sec(y) \big)\Big).
	\end{array}
	\end{equation}
		The least fixpoint in~\eqref{eq:lfpexample} actually defines $r()$ as the transitive closure of $con()$.
		
		Using the well-known Knaster and Tarski theorem, the least fixpoint $\lfp{r(x,y)}{A(r)}$, where $A(r)\defequiv con(x,y)\lor  \exists z \big(con(x,z)\land r(z,y)\big)$,  is equivalent to a disjunction $\displaystyle \bigvee_i A^i(\false)$, where $A^i$ stands for applying $A$ $i$ times. Therefore, the formula~\eqref{eq:lfpexample} can be represented by the conjunction:
		\[
		\begin{array}{ll}
		  \forall y	\Big(& \\
		  &\exists x\big(ex(x)\land A^0(\false)  \big) \implies \big(in(y)\implies sec(y) \big)\land\\
				&\exists x\big(ex(x)\land A^1(\false)  \big) \implies \big(in(y)\implies sec(y) \big)\land\\
				&\ldots\\
				&\exists x\big(ex(x)\land A^i(\false)  \big) \implies \big(in(y)\implies sec(y) \big)\land\\
				&\ldots\\
				\;\;\;\; \Big).
			\end{array}
		\]
		
To compute \sforget{Th(con,r,ex,in,sec);r}, we consider:
\begin{align}
\sforget{Th(con,r,ex,\,&in,sec);r}\equiv\nonumber\\  
    & \forall r\, \Big(\forall x \forall y\Big(\big(con(x,y)\lor\exists z \big(con(x,z)\land 
		r(z,y)\big)\big)\implies r(x,y)\Big)\land\label{eq:sec3}\\
		&\;\;\;\;\;\;\forall y	\Big(\exists x\big(ex(x)\land r(x,y)\big)\implies \big(in(y)\implies sec(y) \big)\Big).\label{eq:sec4}
\end{align}
The quantifier $\forall r$ can be distributed over the conjunction, so one can consider: 
\begin{align}				
& \forall r\, \forall x \forall y\Big(\big(con(x,y)\lor\exists z \big(con(x,z)\land 
r(z,y)\big)\big)\implies r(x,y)\Big)\land\label{eq:sec5}\\
&\forall r\forall y	\Big(\exists x\big(ex(x)\land r(x,y)\big)\implies \big(in(y)\implies sec(y) \big)\Big).\label{eq:sec6}
\end{align}
To eliminate $\forall r$ from~\eqref{eq:sec5}, we transform it to the following equivalent form:
\begin{equation}
     \lneg \exists x \exists y\exists r\, \Big(\big(con(x,y)\lor\exists z \big(con(x,z)\land 
r(z,y)\big)\big)\land\lneg r(x,y)\Big),
\end{equation}
equivalent to:
\begin{equation}\label{eq:zzzz}
    \lneg \exists x \exists y\exists r\, \Big( 
     \underbrace{\forall u\forall w \big(r(u,w)\implies(u\not=x\lor w\not=y))\big)}_{\lneg r(x,y)}\land \big(con(x,y)\lor\exists z \big(con(x,z)\land 
r(z,y)\big)\big)\Big).
\end{equation}
Applying Ackermann's Lemma~\ref{lemma:ack-fo}\eqref{eq:ack-pos-fo} to~\eqref{eq:zzzz} one obtains:
\begin{equation}\label{eq:zzzzfin}
    \lneg  \exists x \exists y \Big( 
      \big(con(x,y)\lor\exists z \big(con(x,z)\land 
(z\not=x\lor y\not=y)\big)\big)\Big),
\end{equation}
equivalent to:
\begin{equation}\label{eq:fin0}
    \forall x \forall y \Big( 
      \big(con(x,y)\implies \forall z \big(con(x,z)\implies z=x\big)\big)\Big),
\end{equation}
which in turn is equivalent to:
\begin{equation}\label{eq:zzzzfin1}
    \forall x \forall z \Big( 
      con(x,z)\implies z=x\Big).
\end{equation}

To eliminate $\forall r$ from~\eqref{eq:sec6}, we transform it to the following equivalent form:
\begin{equation}\label{eq:lasttt}
    \lneg \exists y	\exists r\Big(\exists x\big(ex(x)\land r(x,y)\big)\land  in(y)\land \lneg sec(y) \Big).
\end{equation}
Adding an artificial conjunct $\forall x\forall y\big(r(x,y)\implies \true\big)$, equivalent to \true (to ensure the right syntactic structure), and applying Ackermann's Lemma~\ref{lemma:ack-fo}\eqref{eq:ack-pos-fo} to~\eqref{eq:lasttt}, we obtain:
\begin{equation}\label{eq:lasttt1}
    \forall y\Big(\exists x\big(ex(x)\big)\implies  \big(in(y)\implies sec(y) \Big).
\end{equation}
Combining~\eqref{eq:zzzzfin1} and~\eqref{eq:lasttt1} we obtain that:
\begin{align}
    \sforget{Th(con,r,ex,\,&in,sec);r}\equiv\nonumber\\
    & \forall x \forall z \Big( 
      con(x,z)\implies z=x\Big)\land\\
      & \forall y\Big(\exists x\big(ex(x)\big)\implies  \big(in(y)\implies sec(y) \Big).
\end{align}
That is, to make sure that $Th(con,r,ex,in,sec)$, when $r()$ is forgotten, nodes can  only be connected to themselves and if there is an external node, then every internal node is to be equipped with the security component. Though this guarantees that the security requirements are satisfied, the resulting theory is rather strong. \done
\end{example}

\subsection{Computational Aspects}\label{sec:compasp}

Forgetting is typically applied to finite domain knowledge or belief bases and rule languages. As indicated by the first points of Theorems~\ref{thm:nforget} and~\ref{thm:sforget}, Computing  queries expressed by \nforget{} is NP-complete, and those expressed by \sforget{} is co-NP-complete. On the other hand, the data complexity of first-order queries obtained using Lemma~\ref{lemma:ack-fo} is PTime and LogSpace~\cite{AHV}. Data complexity of fixpoint queries, thus also queries obtained from Lemma~\ref{lemma:fixpoint}, is PTime~\cite{AHV}. Therefore, the approach based on Ackermann's Lemma and its fixpoint extension is computationally friendly.

The following approaches to second-order quantifier elimination that have previously been formulated and implemented are:
\begin{itemize}
    \item using Lemma~\ref{lemma:ack-fo} as a basis for the \dls algorithm~\cite{dls}, which first attempts to transform an arbitrary formula into a form suitable for applying this lemma, and then uses this lemma as a basis for  eliminating quantifiers;
    \item using Lemma~\ref{lemma:fixpoint} as a basis for the \dlsstar algorithm~\cite{dlsstar}, which extends the \dls algorithm to work with formulas that are suitable for an application of the fixpoint lemma.
\end{itemize} 
In fact, all calculations carried out in this paper that involve elimination of second-order quantifiers reflect selected steps used in the \dls or \dlsstar algorithms.

As has been shown in Section~\ref{sec:overhead}, computing propositional equivalents for the weak forgetting operator is often more efficient than computing such equivalents for standard forgetting and for weakest sufficient and  strongest necessary conditions. This reasoning can also be extended to the first-order case as follows, where we consider eliminating the quantifier $\forall r$ from the formula:
\begin{equation}\label{eq:dualcomp}
    \forall r\forall\bar{x}\Big( r(\bar{x}_1)\lor\ldots\lor r(\bar{x}_m)\lor \lneg r(\bar{x}_{m+1})\lor\ldots\lor \lneg r(\bar{x}_n)\lor A(\bar{z}) \Big),
\end{equation}
where $0\leq m\leq n$, $\bar{x}$ contains at least all variables in $\bar{x}_1, \ldots,\bar{x}_n$, and formula $A$ does not contain relation symbol $r$. Formula~\eqref{eq:dualcomp} is equivalent to:
\begin{equation}\label{eq:dualcomp1}
    \forall\bar{x}\lneg \exists r\Big(\lneg r(\bar{x}_1)\land\ldots\land\lneg r(\bar{x}_m)\land  r(\bar{x}_{m+1})\land\ldots\land r(\bar{x}_n)\land \lneg A(\bar{z}) \Big),
\end{equation}
which can be transformed into the equivalent form (e.g., using the  \dls algorithm):
\begin{equation}\label{eq:dualcomp2}
  \forall\bar{x}\lneg \exists r\Big(\forall\bar{y}\big( r(\bar{y})\implies (\bar{y}\not=\bar{x}_1\land\ldots\land\bar{y}\not=\bar{x}_m)\big)\land  r(\bar{x}_{m+1})\land\ldots\land r(\bar{x}_n)\land \lneg A(\bar{z})\Big).
\end{equation}
After applying Ackermann's Lemma~(\ref{lemma:ack-fo}) we obtain the following formula equivalent to~\eqref{eq:dualcomp2}, thus also equivalent to~\eqref{eq:dualcomp}: 
\begin{equation}\label{eq:dualcomp3}
  \forall\bar{x}\lneg \Big( (\bar{x}_{m+1}\not=\bar{x}_1\land\ldots\land\bar{x}_{m+1}\not=\bar{x}_m) \land\ldots\land 
(\bar{x}_{n}\not=\bar{x}_1\land\ldots\land\bar{x}_{n}\not=\bar{x}_m) \land \lneg A(\bar{z})\Big).
\end{equation}
Formula~\eqref{eq:dualcomp3} can be simplified to:
\begin{equation}\label{eq:dualcomp4}
  \forall\bar{x} \Big( \bar{x}_{m+1}=\bar{x}_1\lor\ldots\lor\bar{x}_{m+1}=\bar{x}_m \lor\ldots\lor 
\bar{x}_{n}=\bar{x}_1\lor\ldots\lor\bar{x}_{n}=\bar{x}_m \lor A(\bar{z})\Big).
\end{equation}
That is, the length of the resulting formula~\eqref{eq:dualcomp4} is at most quadratic in the length of the input formula~\eqref{eq:dualcomp}. Given variables $\bar{x}_1,\ldots,\bar{x}_m,\bar{x}_{m+1},\ldots,\bar{x}_n$, one typically can still substantially simplify the result. Notice that in similar cases, the fixpoint lemma is not needed (but may be needed for other shapes of formulas).

When dealing with a conjunction of formulas of the form~\eqref{eq:dualcomp}, the quantifier $\forall r$ can be distributed over conjunction (as, e.g., in~\eqref{eq:testacksc1}), and each conjunct can be processed separately. This provides a sharp contrast in comparison with computing \nforget{}, \wsc{} and \snc{}, where such a~distribution is generally not possible. This is indicated by their syntactic second-order characterizations given by point 1 of Theorem~\ref{thm:nforget}, and formulas~\eqref{eq:so-snc},~\eqref{eq:so-wsc} of Lemma~\ref{lemma:wsc-snc-so}, respectively.

It is worth emphasizing that the problem as to whether a second-order formula is equivalent to a~first-order (or fixpoint) formula, is highly undecidable. So algorithms like \dls, \dlsstar as  well as any other algorithms attempting to do the same quantifier elimination, have to fail for some input formulas. However, pragmatic application has shown that these algorithms work for large classes of formulas (see Section~\ref{sec:relwork} for more a detailed discussion).

\section{Related Work}\label{sec:relwork}

Standard forgetting, \forget(), has been introduced in the foundational paper~\cite{Lin94forgetit}, where model theoretical definitions and analysis of properties of the standard \forget() operator are provided. Its second-order characterization as well as its entailment preserving property, as quoted in Theorem~\ref{thm:forgetit} above, are consequences of this definition. The paper~\cite{Lin94forgetit} opened a~research subarea, summarized, e.g., in~\cite{DitmarschHLM09} or more recently, in~\cite{Eiter-kern-isberner}. In the approach used in the current paper, one begins with an entailment-based, inferential perspective, as expressed by~\eqref{eq:fnc-intuition} and~\eqref{eq:fsc-intuition}. This is directly beneficial for pragmatic application, since this approach leads directly to algorithms and implementations of the dual forgetting operators. In addition to introducing a new weak forgetting operator, standard forgetting is also characterized in this context.

The paper~\cite{DitmarschHLM09} concentrates on {\em introspective} forgetting (``forgetting as becoming
ignorant''). It extends modal epistemic logic with modal operators allowing one to express what is known before and what remains known after forgetting. It therefore essentially deals with necessary conditions, corresponding to standard forgetting. 

In~\cite{Eiter-kern-isberner} two types of forgetting are distinguished: (1) forgetting part of the signature, corresponding to standard forgetting, and (2) forgetting a formula, related to {\em contracting} in the AGM theory of belief change~\cite{AGM85,agm25years}. In the current paper, both dual operators,  weak and strong forgetting, belong to the first type of forgetting. The extension of this work to forgetting formulas is an interesting research direction.

A general framework, covering a~range of belief changes involving forgetting, is proposed in~\cite{BeierleKSBR19}. In particular, a commonsense perspective based on belief change, involving contraction, ignorance introduction, revision/update, (deductive) abstraction at the level of rules, marginalization, etc., is proposed. Another general framework for standard-like forgetting is investigated in~\cite{Delgrande17}, where forgetting is regarded as a~belief change operator, independent of the underlying logic. Forgetting is achieved by reducing a part of a signature in a theory. In that approach, forgetting is specified by the set of logical consequences of the input theory over the reduced language. Though providing very interesting characterizations and contexts of forgetting, the frameworks of~\cite{BeierleKSBR19} and~\cite{Delgrande17} deal with consequences (necessary conditions) of the considered theories and do not deal with sufficient conditions, expressed by the weak forgetting operator in this paper. 

Papers which consider relations between forgetting and weakest sufficient and strongest necessary conditions include~\cite{FengCTL,FengMU}. In the first of these papers, forgetting is applied in the context of strongest necessary and weakest sufficient conditions in Computation Tree Logic (CTL). Axiomatic characterization of forgetting and an algorithm for computing forgetting are provided. In addition, weakest sufficient and strongest necessary conditions are characterized using forgetting (Theorem~9 in~\cite{FengCTL}). The characterization is similar in spirit to the Corollaries~\eqref{eq:corollarysnc} and~\eqref{eq:corollarywsc} in the current paper. In~\cite{FengMU} the results of~\cite{FengCTL} are transferred into the context of $\mu$-calculus. Both in CTL and $\mu$-calculus, typical reasoning problems are intractable. Forgetting in multiagent modal logics, sharing a similar methodology,  has been addressed in~\cite{FangLD19}. Unlike in~\cite{FangLD19,FengCTL,FengMU}, in the current paper, both standard and weak forgetting are considered. The approach isolates broad classes of formulas for which one can compute first-order or fixpoint equivalents of both forgetting operators. This in turn, leads to a~tractable reasoning-by-querying machinery. 

In \cite{ZhangZ10a} a \textit{weak} forgetting operator has been introduced which differs from the weak forgetting operator, \sforget{}, used in this paper, and it has characteristics much more related to the standard forgetting operator, \nforget{}. While standard forgetting is not always first-order expressible, the weak operator of~\cite{ZhangZ10a} has been developed to make sure that the result of forgetting can always be expressed in first-order logic. Weak forgetting in the sense of~\cite{ZhangZ10a} and standard forgetting differ only in the cases where the result of standard forgetting is not first-order expressible. This operator does not coincide with the dual weak forgetting operator \sforget{}. The terminology used in the current paper, reflects the fact that \sforget{} and \nforget{} are dual operators that formally show the relationship of \sforget{} to weakest sufficient conditions and \nforget{} to strongest necessary conditions.

A series of papers~\cite{Del-PintoS19,KoopmannDTS20,ZhaoS17} (see also references there) has addressed forgetting in description logics. Though description logics are strongly related to modal logics~\cite{rijke}, the papers~\cite{Del-PintoS19,KoopmannDTS20,ZhaoS17} are methodologically closer to the approach used in this paper. In~\cite{Del-PintoS19}, forgetting-based abduction is investigated using weakest sufficient and strongest necessary conditions. The idea, inspired by~\cite{dlssnc}, is implemented and experimentally verified using resolution. As in the  approach proposed in the current paper, the papers~\cite{KoopmannDTS20,ZhaoS17} use the Ackermann~\cite{Ack} and fixpoint~\cite{NS} Lemmas, adjusted to the description logic formalism. Though directly dealing with abduction, the approaches in~\cite{Del-PintoS19,KoopmannDTS20,ZhaoS17} concentrate on (valuable and important) experimental research and do not separate the weak forgetting operator from the standard forgetting operator. Forgetting in description logics has also been addressed in~\cite{LutzW11} via  uniform interpolants, where mixed model-theoretic and automata-theoretic approaches are applied. It is shown that computing uniform interpolants in the context of the description logics considered, is generally highly intractable.

Forgetting is particularly useful in rule-based languages, when one simplifies a~belief base to improve querying performance, or protect its parts~\cite{goncalves,WangWZ13,ZHANG2006739}. This is especially useful for Answer Set Programs, where the corresponding entailment tasks, centering around necessary conditions, are typically intractable. In~\cite{WangWZ13} {\em semantic forgetting} is proposed. It preserves skeptical and credulous consequences on unforgotten variables, as well as strong equivalence of ASP programs. It is shown that  computing the forgetting result is intractable even for Horn logic programs.  The paper~\cite{goncalves}, provides a~comprehensive survey of the area as pertains to Answer Set Programs.  

In summary, while the topic of forgetting in knowledge representation has gained considerable attention with numerous publications, the dual weak forgetting operator which is proposed and investigated in this paper, has not been explicitly considered in the literature. Additionally, the fundamental principles of forgetting founded on the entailment-based, inferential perspective, which is the starting point of this paper, is somewhat unique in comparison to other papers, where theoretical results are a consequence of model-theoretical,  or other considerations. This perspective, which results in entailment preservation on a respective sublanguage, naturally leads to the consideration of both standard and weak forgetting operators.

Second-order quantifier elimination is used in this paper as the main logical tool for computing propositional, first-order and fixpoint equivalents of both standard and weak forgetting operators. Although there are several alternative techniques that could be used (see, e.g.,~\cite{gss}), Ackermann-like approaches have been selected for their expressive power and pragmatic application. Combining Ackermann-like lemmas with tautology-preserving formula transformations, as exhibited in this paper, has been shown to be very powerful and useful in many different contexts, including that of computing dual forgetting operators. 

In particular, as shown in~\cite{dls}, the \dls algorithm subsumes most other known techniques that have been developed for computing circumscription. The paper~\cite{Conradie06}, shows that the \dls algorithm covers all Sahlqvist formulas, an important class of formulas used in modal correspondence theory. The fixpoint lemma of~\cite{NS}, has been used originally for application to modal correspondence theory. This in turn has led to the  development of the \dlsstar algorithm for fixpoint computations, which has been used for computing various forms of domain circumscription~\cite{dlsstar}. Both the \dls and \dlsstar algorithms have been shown to be useful in computing weakest sufficient and strongest necessary conditions~\cite{dlssnc}. 
The previously mentioned works~\cite{Del-PintoS19,KoopmannDTS20,ZhaoS17}, in addition to other papers by these authors, have shown that Ackermann-like techniques are also powerful tools for computing forgetting in description logics. The experimental results of these papers also show that such approaches are very efficient and applicable to real-world problems. For other uses of Ackermann-like approaches, including use of the  \dls and \dlsstar algorithms, see~\cite{gss}. 

\section{Conclusions}\label{sec:concl}

This paper provides a general characterization of forgetting founded on an entailment-based, inferential framework. In doing so, an interesting forgetting operator, weak forgetting, is identified which is dual to the standard forgetting operator studied in recent work~\cite{Lin94forgetit}. Due to the entailment-based, inferential framework presented, it is shown how quantifier elimination techniques based on Ackermann's Lemma, using existing algorithms, $DLS$ and $DLS^*$, can be used to compute output of both forgetting operators. Additionally, the tight relationship between weak forgetting and weakest sufficient conditions, and standard forgetting and strongest necessary conditions is characterized in a straightforward manner. 

This paper first approaches the topics using the propositional logic case and then generalizes these results for the first-order and fixpoint logic case, offering new expressivity in the pragmatic use of such operators. Throughout, the paper provides examples justifying the introduction of the weak forgetting operator and also shows how the computational framework is used to derive inferences from application of both forgetting operators.  Similarities and distinctions between related work and work presented in this paper are provided. This work opens up a new way to think about forgetting operators and their pragmatic application in general, by integrating diverse threads of previous research in this area and new results from this paper, in one uniform, formal, entailment-based, inferential framework. 

Based on this approach, it is also shown in Theorems~\ref{thm:nforget}(p.3), \ref{thm:sforget}(p.2), \ref{thm:nforget-fo}(p.3) and \ref{thm:sforget-fo}(p.2) that from the perspective of entailment,  the operators \sforget{} and \nforget{}, generate the formally best possible (maximal) results of forgetting, as expressed by the restricted target language in any application.\footnote{See also Figure~\ref{fig:approx} and the explanations in Section~\ref{sec:best}.}


\begin{thebibliography}{33}
  	\expandafter\ifx\csname natexlab\endcsname\relax\def\natexlab#1{#1}\fi
  	\providecommand{\url}[1]{\texttt{#1}}
  	\providecommand{\href}[2]{#2}
  	\providecommand{\path}[1]{#1}
  	\providecommand{\DOIprefix}{doi:}
  	\providecommand{\ArXivprefix}{arXiv:}
  	\providecommand{\URLprefix}{URL: }
  	\providecommand{\Pubmedprefix}{pmid:}
  	\providecommand{\doi}[1]{\href{http://dx.doi.org/#1}{\path{#1}}}
  	\providecommand{\Pubmed}[1]{\href{pmid:#1}{\path{#1}}}
  	\providecommand{\bibinfo}[2]{#2}
  	\ifx\xfnm\relax \def\xfnm[#1]{\unskip,\space#1}\fi
  	\bibitem[{Abiteboul et~al.(1995)Abiteboul, Hull and Vianu}]{AHV}
  	\bibinfo{author}{Abiteboul, S.}, \bibinfo{author}{Hull, R.},
  	\bibinfo{author}{Vianu, V.}, \bibinfo{year}{1995}.
  	\newblock \bibinfo{title}{Foundations of Databases}.
  	\newblock \bibinfo{publisher}{Addison-Wesley}.
  	\bibitem[{Ackermann(1935)}]{Ack}
  	\bibinfo{author}{Ackermann, W.}, \bibinfo{year}{1935}.
  	\newblock \bibinfo{title}{Untersuchungen \"uber das eliminationsproblem der
  		mathematischen logik}.
  	\newblock \bibinfo{journal}{Mathematische Annalen} \bibinfo{volume}{110},
  	\bibinfo{pages}{390--413}.
  	\bibitem[{Alchourr{\'{o}}n et~al.(1985)Alchourr{\'{o}}n, G{\"{a}}rdenfors and
  		Makinson}]{AGM85}
  	\bibinfo{author}{Alchourr{\'{o}}n, C.}, \bibinfo{author}{G{\"{a}}rdenfors, P.},
  	\bibinfo{author}{Makinson, D.}, \bibinfo{year}{1985}.
  	\newblock \bibinfo{title}{On the logic of theory change: Partial meet
  		contraction and revision functions}.
  	\newblock \bibinfo{journal}{J. Symb. Log.} \bibinfo{volume}{50},
  	\bibinfo{pages}{510--530}.
  	\bibitem[{Areces and de~Rijke(2000)}]{rijke}
  	\bibinfo{author}{Areces, C.}, \bibinfo{author}{de~Rijke, M.},
  	\bibinfo{year}{2000}.
  	\newblock \bibinfo{title}{From description to hybrid logics, and back}, in:
  	\bibinfo{editor}{Wolter, F.}, \bibinfo{editor}{Wansing, H.},
  	\bibinfo{editor}{de~Rijke, M.}, \bibinfo{editor}{Zakharyaschev, M.} (Eds.),
  	\bibinfo{booktitle}{Proc. Advances in Modal Logic 3}, pp.
  	\bibinfo{pages}{17--36}.
  	\bibitem[{Beierle et~al.(2019)Beierle, Kern{-}Isberner, Sauerwald, Bock and
  		Ragni}]{BeierleKSBR19}
  	\bibinfo{author}{Beierle, C.}, \bibinfo{author}{Kern{-}Isberner, G.},
  	\bibinfo{author}{Sauerwald, K.}, \bibinfo{author}{Bock, T.},
  	\bibinfo{author}{Ragni, M.}, \bibinfo{year}{2019}.
  	\newblock \bibinfo{title}{Towards a general framework for kinds of forgetting
  		in common-sense belief management}.
  	\newblock \bibinfo{journal}{K{\"{u}}nstliche Intell.} \bibinfo{volume}{33},
  	\bibinfo{pages}{57--68}.
  	\bibitem[{Conradie(2006)}]{Conradie06}
  	\bibinfo{author}{Conradie, W.}, \bibinfo{year}{2006}.
  	\newblock \bibinfo{title}{On the strength and scope of {DLS}}.
  	\newblock \bibinfo{journal}{J. Appl. Non Class. Logics} \bibinfo{volume}{16},
  	\bibinfo{pages}{279--296}.
  	\bibitem[{van Dalen(2013)}]{vandalen}
  	\bibinfo{author}{van Dalen, D.}, \bibinfo{year}{2013}.
  	\newblock \bibinfo{title}{Logic and Structure}.
  	\newblock Universitext, \bibinfo{publisher}{Springer}.
  	\bibitem[{Del{-}Pinto and Schmidt(2019)}]{Del-PintoS19}
  	\bibinfo{author}{Del{-}Pinto, W.}, \bibinfo{author}{Schmidt, R.},
  	\bibinfo{year}{2019}.
  	\newblock \bibinfo{title}{Abox abduction via forgetting in {ALC}}, in:
  	\bibinfo{booktitle}{The 33rd {AAAI} Conf. on AI}, pp.
  	\bibinfo{pages}{2768--2775}.
  	\bibitem[{Delgrande(2017)}]{Delgrande17}
  	\bibinfo{author}{Delgrande, J.}, \bibinfo{year}{2017}.
  	\newblock \bibinfo{title}{A knowledge level account of forgetting}.
  	\newblock \bibinfo{journal}{J. Artif. Intell. Res.} \bibinfo{volume}{60},
  	\bibinfo{pages}{1165--1213}.
  	\bibitem[{van Ditmarsch et~al.(2009)van Ditmarsch, Herzig, Lang and
  		Marquis}]{DitmarschHLM09}
  	\bibinfo{author}{van Ditmarsch, H.}, \bibinfo{author}{Herzig, A.},
  	\bibinfo{author}{Lang, J.}, \bibinfo{author}{Marquis, P.},
  	\bibinfo{year}{2009}.
  	\newblock \bibinfo{title}{Introspective forgetting}.
  	\newblock \bibinfo{journal}{Synth.} \bibinfo{volume}{169},
  	\bibinfo{pages}{405--423}.
  	\bibitem[{Doherty et~al.(1997)Doherty, {\L}ukaszewicz and Sza{\l}as}]{dls}
  	\bibinfo{author}{Doherty, P.}, \bibinfo{author}{{\L}ukaszewicz, W.},
  	\bibinfo{author}{Sza{\l}as, A.}, \bibinfo{year}{1997}.
  	\newblock \bibinfo{title}{Computing circumscription revisited}.
  	\newblock \bibinfo{journal}{J. Automated Reasoning} \bibinfo{volume}{18},
  	\bibinfo{pages}{297--336}.
  	\bibitem[{Doherty et~al.(1998)Doherty, {\L}ukaszewicz and Sza{\l}as}]{dlsstar}
  	\bibinfo{author}{Doherty, P.}, \bibinfo{author}{{\L}ukaszewicz, W.},
  	\bibinfo{author}{Sza{\l}as, A.}, \bibinfo{year}{1998}.
  	\newblock \bibinfo{title}{General domain circumscription and its effective
  		reductions}.
  	\newblock \bibinfo{journal}{Fundam. Informaticae} \bibinfo{volume}{36},
  	\bibinfo{pages}{23--55}.
  	\bibitem[{Doherty et~al.(2001)Doherty, {\L}ukaszewicz and Sza{\l}as}]{dlssnc}
  	\bibinfo{author}{Doherty, P.}, \bibinfo{author}{{\L}ukaszewicz, W.},
  	\bibinfo{author}{Sza{\l}as, A.}, \bibinfo{year}{2001}.
  	\newblock \bibinfo{title}{Computing strongest necessary and weakest sufficient
  		conditions of first-order formulas}, in: \bibinfo{editor}{Nebel, B.} (Ed.),
  	\bibinfo{booktitle}{Proc. 17th {IJCAI}}, pp. \bibinfo{pages}{145--154}.
  	\bibitem[{Eiter and Kern{-}Isberner(2019)}]{Eiter-kern-isberner}
  	\bibinfo{author}{Eiter, T.}, \bibinfo{author}{Kern{-}Isberner, G.},
  	\bibinfo{year}{2019}.
  	\newblock \bibinfo{title}{A brief survey on forgetting from a knowledge
  		representation and reasoning perspective}.
  	\newblock \bibinfo{journal}{K\"unstliche Intell.} \bibinfo{volume}{33},
  	\bibinfo{pages}{9--33}.
  	\bibitem[{Fang et~al.(2019)Fang, Liu and van Ditmarsch}]{FangLD19}
  	\bibinfo{author}{Fang, L.}, \bibinfo{author}{Liu, Y.}, \bibinfo{author}{van
  		Ditmarsch, H.}, \bibinfo{year}{2019}.
  	\newblock \bibinfo{title}{Forgetting in multi-agent modal logics}.
  	\newblock \bibinfo{journal}{Artif. Intell.} \bibinfo{volume}{266},
  	\bibinfo{pages}{51--80}.
  	\bibitem[{Feng et~al.(2022)Feng, Acar, Wang, Liu, Schlobach and Ding}]{FengCTL}
  	\bibinfo{author}{Feng, R.}, \bibinfo{author}{Acar, E.}, \bibinfo{author}{Wang,
  		Y.}, \bibinfo{author}{Liu, W.}, \bibinfo{author}{Schlobach, S.},
  	\bibinfo{author}{Ding, W.}, \bibinfo{year}{2022}.
  	\newblock \bibinfo{title}{Computing sufficient and necessary conditions in
  		{CTL:} {A} forgetting approach}.
  	\newblock \bibinfo{journal}{Inf. Sci.} \bibinfo{volume}{616},
  	\bibinfo{pages}{474--504}.
  	\bibitem[{Feng et~al.(2023)Feng, Wang, Qian, Yang and Chen}]{FengMU}
  	\bibinfo{author}{Feng, R.}, \bibinfo{author}{Wang, Y.}, \bibinfo{author}{Qian,
  		R.}, \bibinfo{author}{Yang, L.}, \bibinfo{author}{Chen, P.},
  	\bibinfo{year}{2023}.
  	\newblock \bibinfo{title}{Knowledge forgetting in propositional
  		$\mu$-calculus}.
  	\newblock \bibinfo{journal}{Ann. Math. Artif. Intell.} ,
  	\bibinfo{pages}{1--43}.
  	\bibitem[{Ferm\'e and Hansson(2011)}]{agm25years}
  	\bibinfo{author}{Ferm\'e, E.}, \bibinfo{author}{Hansson, S.O.},
  	\bibinfo{year}{2011}.
  	\newblock \bibinfo{title}{{AGM} 25 years: Twenty-five years of research in
  		belief change}.
  	\newblock \bibinfo{journal}{J. of Philosophical Logic} \bibinfo{volume}{40},
  	\bibinfo{pages}{295--331}.
  	\bibitem[{Gabbay et~al.(2008)Gabbay, Schmidt and Sza{\l}as}]{gss}
  	\bibinfo{author}{Gabbay, D.}, \bibinfo{author}{Schmidt, R.},
  	\bibinfo{author}{Sza{\l}as, A.}, \bibinfo{year}{2008}.
  	\newblock \bibinfo{title}{Second-Order Quantifier Elimination. Foundations,
  		Computational Aspects and Applications}. volume~\bibinfo{volume}{12} of
  	\textit{\bibinfo{series}{Studies in Logic}}.
  	\newblock \bibinfo{publisher}{College Pub.}
  	\bibitem[{Gon{\c{c}}alves et~al.(2021)Gon{\c{c}}alves, Knorr and
  		Leite}]{goncalves}
  	\bibinfo{author}{Gon{\c{c}}alves, R.}, \bibinfo{author}{Knorr, M.},
  	\bibinfo{author}{Leite, J.}, \bibinfo{year}{2021}.
  	\newblock \bibinfo{title}{Forgetting in answer set programming - {A} survey}.
  	\newblock \bibinfo{journal}{Theory and Practice of Logic Programming} ,
  	\bibinfo{pages}{1--43}.
  	\bibitem[{Hinman(2005)}]{hinman}
  	\bibinfo{author}{Hinman, P.}, \bibinfo{year}{2005}.
  	\newblock \bibinfo{title}{Fundamentals of Mathematical Logic}.
  	\newblock \bibinfo{publisher}{Taylor \& Francis}.
  	\bibitem[{Koopmann et~al.(2020)Koopmann, Del{-}Pinto, Tourret and
  		Schmidt}]{KoopmannDTS20}
  	\bibinfo{author}{Koopmann, P.}, \bibinfo{author}{Del{-}Pinto, W.},
  	\bibinfo{author}{Tourret, S.}, \bibinfo{author}{Schmidt, R.},
  	\bibinfo{year}{2020}.
  	\newblock \bibinfo{title}{Signature-based abduction for expressive description
  		logics}, in: \bibinfo{editor}{Calvanese, D.}, \bibinfo{editor}{Erdem, E.},
  	\bibinfo{editor}{Thielscher, M.} (Eds.), \bibinfo{booktitle}{Proc. 17th Int.
  		Conf. {KR}'2020}, pp. \bibinfo{pages}{592--602}.
  	\bibitem[{Lang et~al.(2003)Lang, Liberatore and Marquis}]{LangLM03}
  	\bibinfo{author}{Lang, J.}, \bibinfo{author}{Liberatore, P.},
  	\bibinfo{author}{Marquis, P.}, \bibinfo{year}{2003}.
  	\newblock \bibinfo{title}{Propositional independence: Formula-variable
  		independence and forgetting}.
  	\newblock \bibinfo{journal}{J. Artif. Intell. Res.} \bibinfo{volume}{18},
  	\bibinfo{pages}{391--443}.
  	\bibitem[{Lang and Marquis(2010)}]{LangM10}
  	\bibinfo{author}{Lang, J.}, \bibinfo{author}{Marquis, P.},
  	\bibinfo{year}{2010}.
  	\newblock \bibinfo{title}{Reasoning under inconsistency: {A} forgetting-based
  		approach}.
  	\newblock \bibinfo{journal}{Artif. Intell.} \bibinfo{volume}{174},
  	\bibinfo{pages}{799--823}.
  	\bibitem[{Lin(2000)}]{Lin}
  	\bibinfo{author}{Lin, F.}, \bibinfo{year}{2000}.
  	\newblock \bibinfo{title}{On strongest necessary and weakest sufficient
  		conditions}, in: \bibinfo{editor}{Cohn, A.}, \bibinfo{editor}{Giunchiglia,
  		F.}, \bibinfo{editor}{Selman, B.} (Eds.), \bibinfo{booktitle}{Proc. 7th Int.
  		Conf. {KR}'2000}, pp. \bibinfo{pages}{167--175}.
  	\bibitem[{Lin and Reiter(1994)}]{Lin94forgetit}
  	\bibinfo{author}{Lin, F.}, \bibinfo{author}{Reiter, R.}, \bibinfo{year}{1994}.
  	\newblock \bibinfo{title}{Forget it!}, in: \bibinfo{booktitle}{In Proc. of the
  		{AAAI} Fall Symp. on Relevance}, pp. \bibinfo{pages}{154--159}.
  	\bibitem[{Lutz and Wolter(2011)}]{LutzW11}
  	\bibinfo{author}{Lutz, C.}, \bibinfo{author}{Wolter, F.}, \bibinfo{year}{2011}.
  	\newblock \bibinfo{title}{Foundations for uniform interpolation and forgetting
  		in expressive description logics}, in: \bibinfo{editor}{Walsh, T.} (Ed.),
  	\bibinfo{booktitle}{Proc. {IJCAI}'11}, pp. \bibinfo{pages}{989--995}.
  	\bibitem[{Nonnengart and Sza{\l}as(1998)}]{NS}
  	\bibinfo{author}{Nonnengart, A.}, \bibinfo{author}{Sza{\l}as, A.},
  	\bibinfo{year}{1998}.
  	\newblock \bibinfo{title}{A fixpoint approach to second-order quantifier
  		elimination with applications to correspondence theory}, in:
  	\bibinfo{editor}{Or{\l}owska, E.} (Ed.), \bibinfo{booktitle}{Logic at Work:
  		Essays Dedicated to the Memory of Helena Rasiowa},
  	\bibinfo{publisher}{Springer Physica-Verlag}. pp. \bibinfo{pages}{307--328}.
  	\bibitem[{Peppas(2008)}]{Peppas08}
  	\bibinfo{author}{Peppas, P.}, \bibinfo{year}{2008}.
  	\newblock \bibinfo{title}{Belief revision}, in: \bibinfo{editor}{van Harmelen,
  		F.}, \bibinfo{editor}{Lifschitz, V.}, \bibinfo{editor}{Porter, B.} (Eds.),
  	\bibinfo{booktitle}{Handbook of KR}. \bibinfo{publisher}{Elsevier}, pp.
  	\bibinfo{pages}{317--359}.
  	\bibitem[{Wang et~al.(2013)Wang, Wang and Zhang}]{WangWZ13}
  	\bibinfo{author}{Wang, Y.}, \bibinfo{author}{Wang, K.}, \bibinfo{author}{Zhang,
  		M.}, \bibinfo{year}{2013}.
  	\newblock \bibinfo{title}{Forgetting for {A}nswer {S}et {P}rograms revisited},
  	in: \bibinfo{editor}{Rossi, F.} (Ed.), \bibinfo{booktitle}{Proc.
  		{IJCAI}'2013}, pp. \bibinfo{pages}{1162--1168}.
  	\bibitem[{Zhang and Foo(2006)}]{ZHANG2006739}
  	\bibinfo{author}{Zhang, Y.}, \bibinfo{author}{Foo, N.}, \bibinfo{year}{2006}.
  	\newblock \bibinfo{title}{Solving logic program conflict through strong and
  		weak forgettings}.
  	\newblock \bibinfo{journal}{Artificial Intelligence} \bibinfo{volume}{170},
  	\bibinfo{pages}{739--778}.
  	\bibitem[{Zhang and Zhou(2010)}]{ZhangZ10a}
  	\bibinfo{author}{Zhang, Y.}, \bibinfo{author}{Zhou, Y.}, \bibinfo{year}{2010}.
  	\newblock \bibinfo{title}{Forgetting revisited}, in: \bibinfo{editor}{Lin, F.},
  	\bibinfo{editor}{Sattler, U.}, \bibinfo{editor}{Truszczynski, M.} (Eds.),
  	\bibinfo{booktitle}{Proc. 12th Int. Conf. {KR}'2010}, pp.
  	\bibinfo{pages}{602--604}.
  	\bibitem[{Zhao and Schmidt(2017)}]{ZhaoS17}
  	\bibinfo{author}{Zhao, Y.}, \bibinfo{author}{Schmidt, R.},
  	\bibinfo{year}{2017}.
  	\newblock \bibinfo{title}{Role forgetting for {ALCOQH} (universal
  		role)-ontologies using an {A}ckermann-based approach}, in:
  	\bibinfo{editor}{Sierra, C.} (Ed.), \bibinfo{booktitle}{Proc. {IJCAI}'17},
  	pp. \bibinfo{pages}{1354--1361}.
  	
  \end{thebibliography}
 \end{document}